\documentclass[sn-nature]{sn-jnl}

\usepackage{graphicx}%
\usepackage{multirow}%
\usepackage{amsmath,amssymb,amsfonts}%
\usepackage{amsthm}%
\usepackage{mathrsfs}%
\usepackage[title]{appendix}%
\usepackage{xcolor}%
\usepackage{textcomp}%
\usepackage{manyfoot}%
\usepackage{booktabs}%
\usepackage{algorithm}%
\usepackage{algorithmicx}%
\usepackage{algpseudocode}%
\usepackage{listings}%

\theoremstyle{thmstyleone}%
\theoremstyle{thmstyletwo}%

\theoremstyle{thmstylethree}%

\begin{document}

\title[Article Title]{Implicit Neural Image Field for Biological Microscopy Image Compression}


\author[1]{\fnm{Gaole} \sur{Dai}}

\author[1]{\fnm{Cheng-Ching} \sur{Tseng}}\equalcont{Equal contribution.}

\author[1]{\fnm{Qingpo} \sur{Wuwu}}\equalcont{Equal contribution.}

\author[1]{\fnm{Rongyu} \sur{Zhang}}\equalcont{Equal contribution.}

\author[1]{\fnm{Shaokang} \sur{Wang}}\equalcont{Equal contribution.}

\author[1]{\fnm{Ming} \sur{Lu}}

\author[1]{\fnm{Tiejun} \sur{Huang}}

\author[2]{\fnm{Yu} \sur{Zhou}}

\author[3]{\fnm{Ali Ata} \sur{Tuz}}

\author[2,3]{\fnm{Matthias} \sur{Gunzer}}

\author*[2]{\fnm{Jianxu} \sur{Chen}}\email{jianxu.chen@isas.de}\equalsup{Equal Supervision. Authors are permitted to list their name last in their CVs}

\author*[1]{\fnm{Shanghang} \sur{Zhang}}\email{shanghang@pku.edu.cn}\equalsup{Equal Supervision. Authors are permitted to list their name last in their CVs}

\affil[1]{\orgdiv{National Key Laboratory for Multimedia Information Processing}, \orgname{Peking University}, \orgaddress{\city{Beijing}, \postcode{100871}, \country{China}}}

\affil[2]{\orgname{Leibniz-Institut für Analytische Wissenschaften – ISAS – e.V.}, \orgaddress{\street{Bunsen-Kirchhoff-Str. 11 }, \city{Dortmund}, \postcode{44139}, \country{Germany}}}

\affil[3]{\orgdiv{Institute for Experimental Immunology and Imaging}, \orgname{University of Duisburg-Essen}, \orgaddress{\street{Universitätsstr. 2}, \city{Essen}, \postcode{45141}, \country{Germany}}}


\abstract{The rapid pace of innovation in biological microscopy imaging has led to large images, putting pressure on data storage and impeding efficient sharing, management, and visualization. This necessitates the development of efficient compression solutions. Traditional CODEC methods struggle to adapt to the diverse bioimaging data and often suffer from sub-optimal compression. In this study, we propose an adaptive compression workflow based on Implicit Neural Representation (INR). This approach permits application-specific compression objectives, capable of compressing images of any shape and arbitrary pixel-wise decompression. We demonstrated on a wide range of microscopy images from real applications that our workflow not only achieved high, controllable compression ratios (e.g., 512x) but also preserved detailed information critical for downstream analysis.}




\maketitle

\section{Introduction}\label{sec1}
\begin{figure}[htb]
    \centering
    \includegraphics[width=1\textwidth]{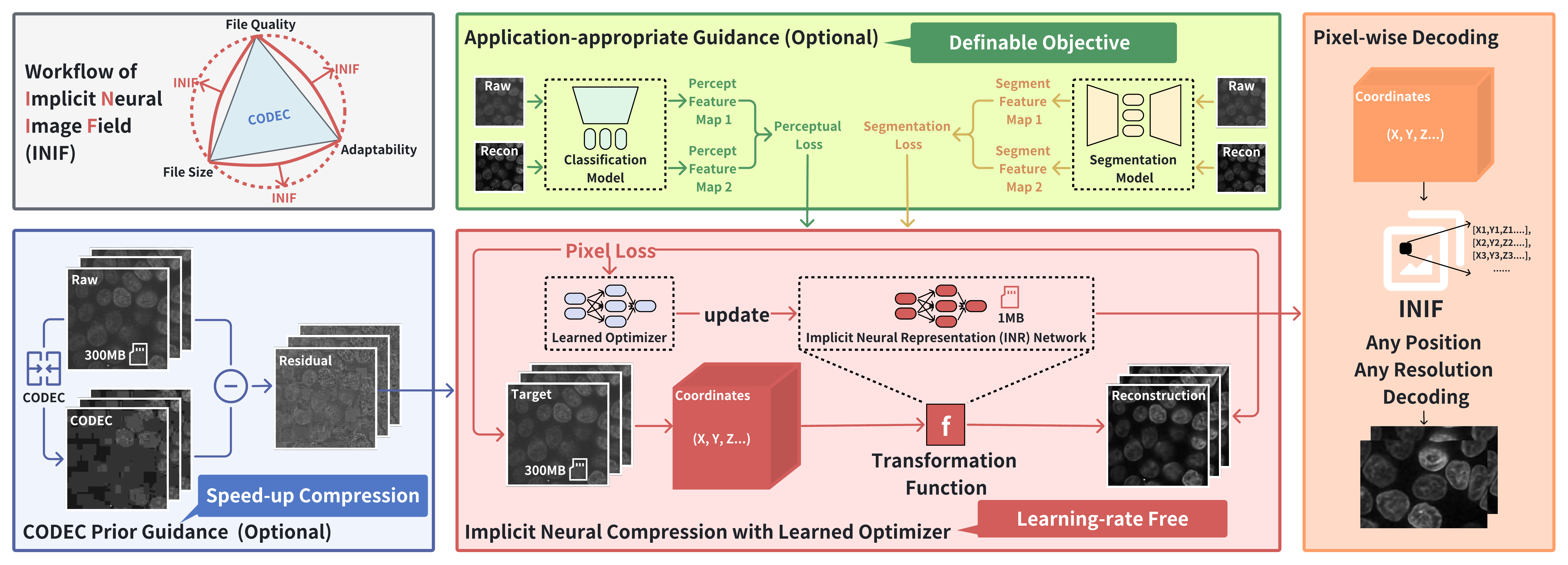}
\caption{\textbf{The INIF compression/decompression workflow.} The general training process begins by generating coordinates that have the same shape as the target compression images. These coordinates are then fed into the INR model to transform their corresponding pixel values. The target images are used as the ground truth for calculating the per-pixel similarity loss, which is optimized using a learned optimizer to update the INR model iteratively. (red panel) Additional guidance, such as application-specific loss (green panel) and CODEC prior (blue panel), can be incorporated when necessary. The final storage content is a unique INIF file containing the weights of the INR model and the metadata. The INIF file can be decoded in multiple flexible ways, e.g., a single Z-slice, a collection of region-of-interest, a low-resolution version (only decoding every 4 pixels along XY), etc., with GPU support (orange panel).}
    \label{fig1}
\end{figure}

Microscopic imaging is essential in modern biological research, not only for observational discovery but also for statistical analyses and mathematical modelling. The past three decades have witnessed significant advancements in imaging techniques and analysis methods in tandem. But, this progress has also presented a new challenge: the efficient storage and computational management of this massively increasing volume of microscopy data. While recent efforts in metadata standardization~\cite{hammer2021towards}, data management tools~\cite{allan2012omero} and next-generation file formats (NGFF)~\cite{moore2021ome} have alleviated some challenges, the sheer size of the microscopy data remains a concern.

To address this issue, we revisit a classic technique - image compression, which is not systematically optimized for microscopy data yet. While commercial codecs, like JPEG2000~\cite{christopoulos2000jpeg2000} and high-efficiency video coding (HEVC)~\cite{sze2014high}, offer some utility, they are not primarily designed for the unique characteristics of bioimages. For example, microscopy images could be of up to five dimensions, X(length), Y(width), Z(depth), C(channel), and T(time), which could offer much more contextual information and therefore have much more potential redundancy to be compressed than standard RGB images or videos. Also, special optical properties for different light microscopes, such as the use of narrow-band filters in fluorescence microscopy or the polarized light source in differential interference contrast (DIC) microscopy, may cause the compression schemes designed for nature images yielding sub-optimal results for bioimages.

Recent progress in artificial intelligence (AI) has yielded promising new compression techniques, one of which is based on implicit neural representations (INR)~\cite{sitzmann2020implicit,molaei2023implicit}. Briefly, INR learns a sample-specific function using an artificial neural network (ANN), usually a small multi-layer perceptron (MLP)~\cite{taud2018multilayer}, to map the multi-dimensional coordinates of an image to its corresponding pixel values. Only the weights of the ANN and dimensional information are stored, enabling on-demand reconstruction of the full image or any specific regions (even disconnected). The flip side of such sample-specific learning is that it might require considerable time to compress one single image while the performance still largely depends on the tuning of certain training hyper-parameters, e.g., learning rate.

In this work, we propose a new microscopy image compression paradigm, called implicit neural image field (INIF). INIF builds upon the effectiveness and flexibility of INR and addresses common bottlenecks in existing INR-based compression algorithms by adopting a learned optimizer~\cite{metz2022velo} and leveraging common codecs for faster compression. More importantly, following the spirit of application-appropriate validation~\cite{chen_when_2023}, INIF integrates application-specific guidance for improved compression quality and trustworthiness. We believe INIF could open new avenues for microscopy image data management and stimulate further development from the bioimaging community.

\section{Results}\label{sec2}

The overall workflow of our INIF microscopy image compression framework is illustrated in Figure~\ref{fig1}. The compression process is to train the INR network. The INR network is usually a small multi-layer perception (MLP) with several linear layers, while the exact model architecture is decided according to the target compression ratio and the shape information (see section~\ref{sec5-1}). The core of this training process starts with generating coordinate grids with identical shapes as the corresponding multi-dimensional microscopy data. These coordinates are utilized as inputs and fed into the INR network to transform their respective pixel values. The raw image serves as ground truth for calibrating the similarity between the reconstructed image and the raw image with a similarity loss (e.g. mean square error), subsequently optimized by a learned optimizer~\cite{metz2022velo} (i.e.,  no need to set learning ratio or other optimization hyper-parameters). 

Besides the core, there are two major optional components in INIF. The first one is application-appropriate guidance, by adding additional loss functions besides the default per-pixel similarity loss. In section~\ref{sec2.2}, we will demonstrate two examples of additional application-appropriate guidance, one with the application objective of maintaining downstream segmentation quality, and the other with the application objective of increasing the model robustness on noisy images acquired with very low laser power. The second optional component is the utilization of a classic codec as prior guidance for fast compression. This component was motivated by the fact that classic codec has clear advantages in terms of compression speed. Then, we can use the INR network to compress only the residual after codec compression (i.e., the difference between the raw image and the codec-compressed image). This residual compression mechanism can significantly reduce the compression time since most of the information is compressed by the fast codec and the amount of information that the INR network needs to learn is greatly reduced. One example will be demonstrated in section \ref{sec2.3}. 

The compression performance was compared against a representative commercial codec, HEVC, and an existing INR-based compression baseline method, SIREN~\cite{sitzmann2020implicit}, with visual qualitative demonstration and additional similarity metrics. It is critical to know that whether the compression quality is sufficient depends on the downstream tasks, and users could confirm the validity via application-appropriate validations~\cite{chen_when_2023} before deployment.

After the compression process, i.e., training the INR network, the model weights and shape information (with additional compressed bitstream by a classic codec when residual compression is used) are the only bytes to store on disk. In order to decompress, or reconstruct the image, a grid of the full image size or a collection of certain grid points can be fed into the INR network to retrieve the full image or any subparts of the full image. As illustrated in Figure \ref{fig1}, different from classic codec, like JPEG, PNG, or TIFF, where the full image is always decoded as a whole, or more sophisticated file formats, like Zarr-based formats~\cite{abernathey2021cloud}, where information can be retrieved chunk by chunk or only retrieve a pre-calculated low-resolution thumbnail, INIF offers the ultimate flexibility to efficiently retrieve any region of interest or even sub-sampled preview (e.g., retrieve one pixel every four grid points along X dimension for 4x downsampling along X, similar for Y and Z dimensions). Such flexibility in decompression holds great potential to be integrated into large-scale analysis, browser-based big image visualization, etc..

\subsection{Compression of Microscopy Images of Different Dimensionalities}\label{sec2.1}

\begin{figure}[htb]
    \centering
    \includegraphics[width=1\textwidth]{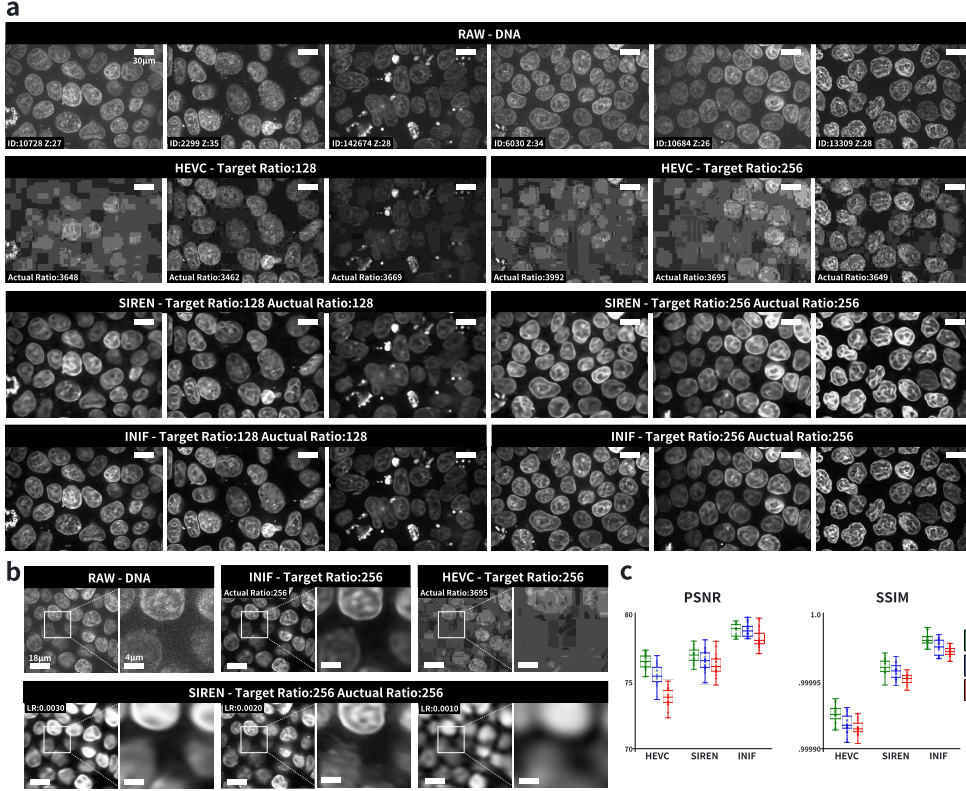}
    \caption{\textbf{Comparison of INIF, HEVC and SIREN on 3D DNA-stained hiPSC data.} \textbf{a,} The visualization results demonstrate that HEVC exhibits noticeable blocking artifacts (row 2). SIREN successfully captures the overall contour of the nucleus but lacks a detailed reconstruction of its edges and inner contents (row 3). In contrast, INIF effectively maintains clear visibility of cellular contours and intranuclear material distribution even under high compression ratios, such as 128x and 256x (row 4). \textbf{b,} SIREN is very sensitive to the setting of the learning rate (row 2), while INIF generates sharper compressed images without learning rate optimization (row 1, middle). \textbf{c,} Quantitative assessment of restoration error using PSNR (left) and SSIM (right) metrics (higher is better) across different compression ratios (128x, 256x, and 512x).}
    \label{fig2}
\end{figure}

\textbf{Compression of Volumetric Microscopy Images (ZYX)}: Volumetric light microscopy images are widely used in biology considering the 3D nature of biological samples. Common 2D image compression codecs without taking the lateral context into account could be problematic, while the time dimension for common video compression codecs has a different assumption than the lateral dimension of 3D microscopy due to special optical properties. We demonstrate the effectiveness of INIF compression on a 100x high-resolution 3D confocal spinning disk microscopy image of human induced pluripotent stem cells (hiPSC). The images were from the public hiPSC image dataset~\cite{viana2023integrated} released by the Allen Institute for Cell Science, consisting of multi-channel 3D images acquired from 25 different cell lines and various microscope imaging pipelines. We took the DNA dye channel to test the compression performance, with 25 randomly selected images, one from each cell line to cover all major acquisition variations. Example results are visualized in Figure~\ref{fig2}-a. We can observe that images compressed with HEVC exhibit significant blocking artifacts, which greatly hinder the discernment of nucleoli and chromatin within cell nuclei. Furthermore, even with meticulous adjustment of hyper-parameters such as bitrate~\cite{power2017guide}, constant rate factor (CRF)~\cite{bienik2017impact}, and adaptive quantization (AQ) mode~\cite{xiang2018novel}, it is unable to control the compression rate throughout the HEVC experiment. In contrast, our proposed INIF method consistently maintains clear visibility of cellular contours and retains considerable textural details under controllable high compression ratios of 128 and 256-fold. On the other hand, unlike SIREN which is very sensitive to training hyper-parameters and therefore requires tuning with various learning rates for improved quality (see \ref{fig2}-b), INIF can achieve compression of better quality without any learning rate tuning effort. To quantify compression results, we access the restoration error by comparing decompressed images with ground-truth images across three distinct compression ratios (128x, 256x, 512x, see \ref{fig2}-c). We observed that INIF outperformed HEVC and SIREN in both the Peak Signal-to-Noise Ratio (PSNR) and Structural Similarity Index (SSIM)~\cite{hore2010image} by a large margin.

\begin{figure}[htb]
    \centering
    \includegraphics[width=0.95\textwidth]{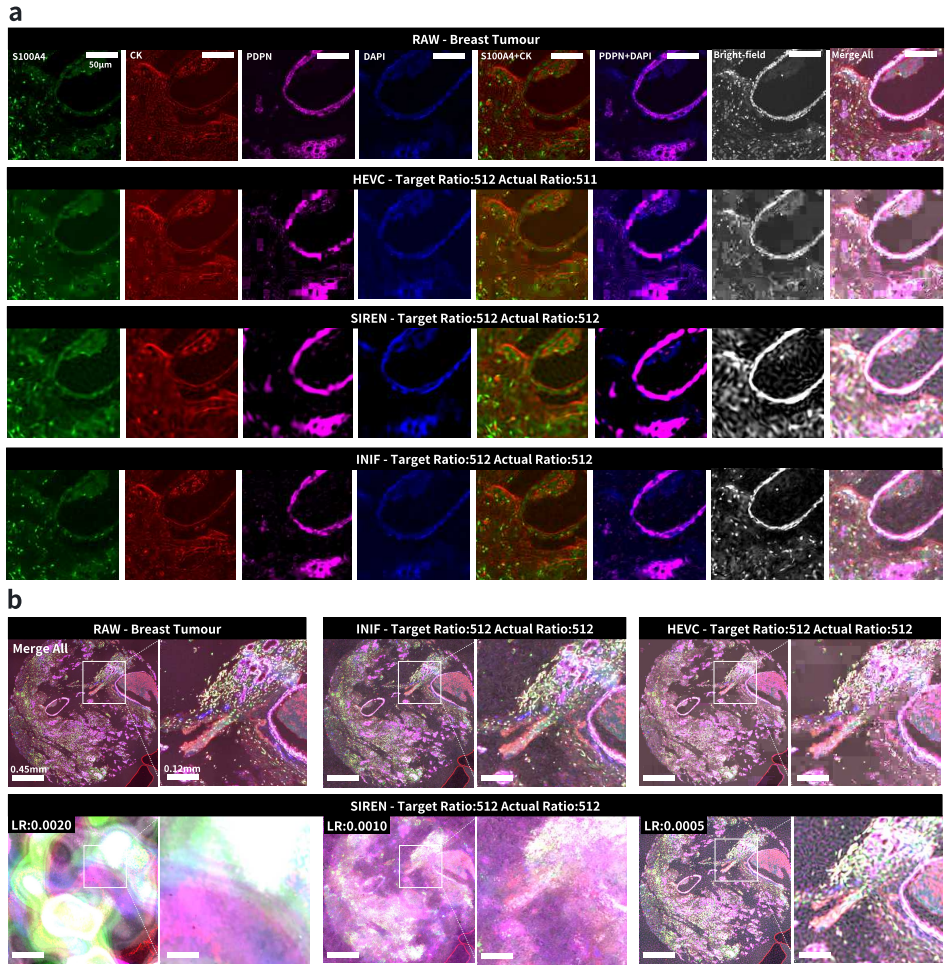}
    \caption{\textbf{Comparison of INIF, HEVC and SIREN on multichannel breast tumour data.} \textbf{a,} HEVC yields unsatisfactory results with noticeable blocking artifacts, especially visible in the merged image (row 2). SIREN  exhibits issues of over-smoothing, compromising the quality of merged outcomes (row 3). On the other hand, INIF demonstrates more satisfactory results in both individual and merged outputs than the others (row 4). \textbf{b,} SIREN exhibits sensitivity to the learning rate, leading to undesired results with minor changes (bottom row).}
    \label{fig3}
\end{figure}

\textbf{Compression of Multiplexed Microscopy Images (CYX)}: Multiplex microscopy, e.g., multiplexed immunofluorescence (MxIF)~\cite{francisco2020multiplex}, enables simultaneous detection and co-localization analysis of multiple markers, making it an indispensable tool for uncovering the heterogeneity in biological studies. Here, the additional information along the channel dimension, usually more than three (compared to three channels in RGB images), makes common codec sub-optimal. We verified the performance of INIF on a public dataset, 1-TNBC (image data resource (IDR) project ID 2801)~\cite{friedman2020cancer}. It comprises MxIF imaging of breast tumour tissues from triple-negative breast cancer (TNBC) patients. Each sample includes five channels: S100A4-Opal 520, CK-Opal 570, PDPN-Opal 650, DAPI, and bright-field with 3506x3506 pixels along XY. 
Specifically, S100A4 and PDPN are proteins that exhibit high expression in cancer-associated fibroblasts (CAFs), while CK stains the epithelial regions and DAPI stains adenine-thymine-rich regions in DNA. By merging the S100A4 or PDPN channel with the CK or DAPI channel, co-localization analysis within the corresponding micro-environment can be performed~\cite{friedman2020cancer}. The objective, in this case, is to preserve CAF heterogeneity after compression. The results are shown in Figure~\ref{fig3}. Upon visualization, it is evident that HEVC produced unsatisfactory results, where the presence of blocking artifacts significantly affects the final quality. As in the volumetric microscopy images example, we observed a similar issue where SIREN exhibits sensitivity to the learning rate; even minor changes in the learning rate can lead to corrupted results (Figure~\ref{fig3}-b).

\begin{figure}[htb]
    \centering
    \includegraphics[width=\textwidth]{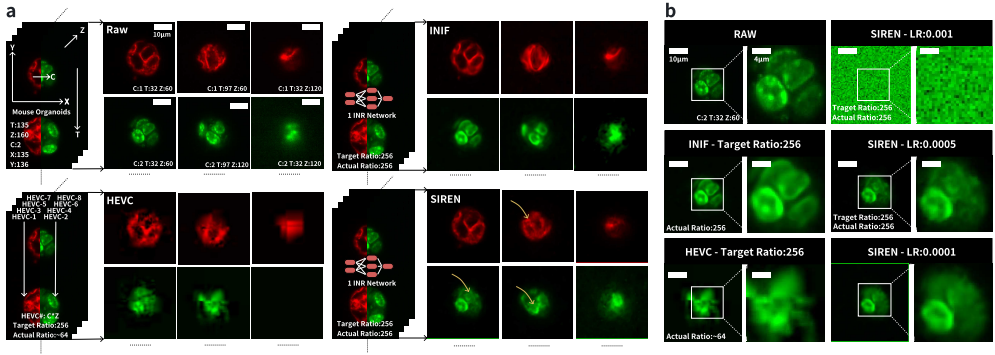}
    \caption{\textbf{Comparison of INIF, HEVC and SIREN on 5D mouse organoid data.} \textbf{a,} HEVC struggles to control the final compression ratio and produces decompressed images with merged cell boundaries, obscuring the precise localization and three-dimensional structure of cells (bottom left).
    Although both SIREN (top right) and INIF (bottom right) maintain a consistent compression ratio, only INIF preserves exceptional detail (errors in the SIREN are indicated by yellow arrows). This enables clear differentiation of individual cell outlines and facilitates vivid observation of dynamic changes in cells over time. \textbf{b,} The SIREN model still heavily relies on selecting an appropriate learning rate range for training, as the performance is significantly affected by distributions that fall outside of this range.}
    \label{fig4}
\end{figure}

\textbf{Compression of Multi-channel Volumetric Timelapse Microscopy Images (TCZYX)}: Microscopy images can be up to 5D, which makes commercial codecs not directly applicable to the full image. We could use such codecs to independently compress the data slice-by-slice, which completely loses the contextual information. 
Consequently, such fragmentation leads to a noticeable deterioration in the quality of the resulting compressed images. We demonstrate the applicability of INIF in 5D microscopy images, with a public dataset of mouse organoids imaged with dual oblique plane microscopy (dOPM)~\cite{Smith_Sparks_Almagro_Behrens_Dunsby_Salbreux_Chaigne_2023} over time. The 5D movies showcased the growth of mouse organoids. The primary objective of this dataset is to discern the three-dimensional dynamics in the membrane and DNA channels over time during mitosis. It is crucial to maintain the cellular shape and count even after compression. Upon analyzing the results obtained under a target compression ratio of 256 (Figure~\ref{fig4}), we have observed that controlling the final compression ratio of HEVC is challenging due to the need to align numerous individual compression processes. The boundaries between cells in decompressed images using HEVC have become indistinguishable, thereby obscuring the precise localization and three-dimensional structure of cells within the organoid—the results from SIREN exhibit visible alteration on granular structures in membrane and DNA. In contrast, INIF not only maintains a stable compression ratio but also preserves an exceptional level of detail. This preservation enables clear differentiation of individual cell outlines using just one single INR network (Figure~\ref{fig4}).  

\subsection{Application-appropriate guidance with adjustable compression objective}\label{sec2.2}

\begin{figure}[htbp]
    \centering
    \includegraphics[width=0.85\textwidth]{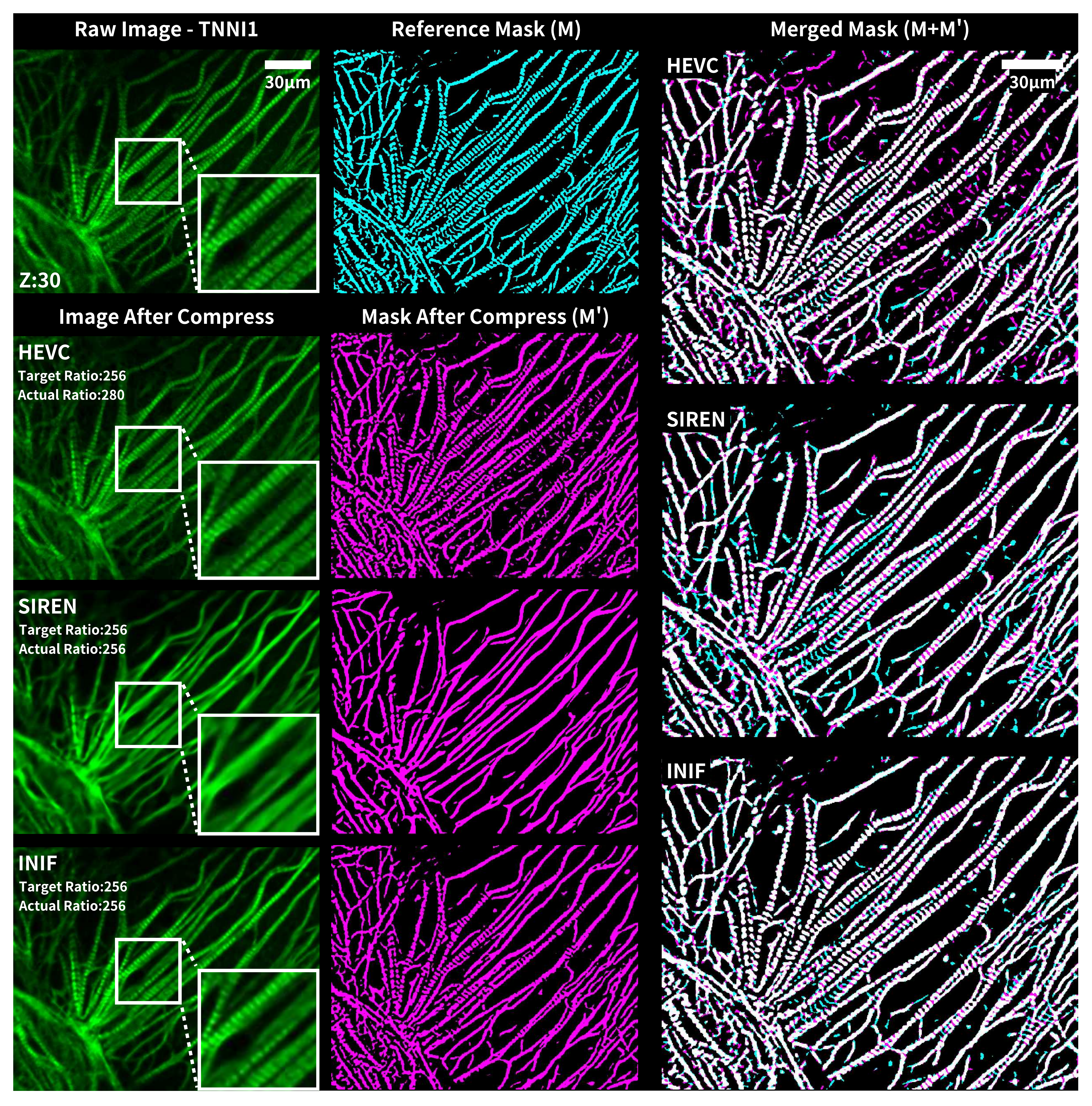}
    \caption{\textbf{INIF with segmentation guidance on TNNI1-stained data.} INIF effectively retains smaller components such as sarcomeres with section-by-section structures (row 4 column 1). The optimal outcome should demonstrate the precise overlap between the magenta (mask after compression) and cyan masks (reference mask), resulting in a binary mask of black-white (true negative and true positive). Sub-optimal outcomes would exhibit two colours: cyan values indicating missed segmentation by excluding existing areas with decompressed data (false negative), while magenta values representing incorrect segmentation by including non-existing areas with decompressed data (false positive). The segmentation results show that INIF achieves balanced outcomes compared to HEVC and SIREN.}
    \label{fig5}
\end{figure}

The compression process in INIF can be extended with additional guidance according to specific downstream tasks or requirements. The guidance is achieved via optimization with task-specific loss. In this section, we demonstrate two different application-appropriate guidance examples, one example where we aim to optimize the compression to maintain the accuracy in a downstream segmentation task, and another example where we aim to reduce the noisy signals in the process of compression.



Automatic image segmentation is a common downstream task in quantitative biology studies. We hypothesize that making the compression process aware of the specific downstream segmentation task could help retain the critical, maybe very subtle, information for downstream segmentation (not necessarily critical for visualization). For demonstration, we tested the compression on a sample 3D confocal microscopy (100x) of Troponin I Type 1 (TNNI1) in cardiomyocytes, a sample image released in ~\cite{chen2018allen}. TNNI1 displays a distinct filamentous morphology with striped patterns. As shown in Figure \ref{fig5}, across various decompressed outcomes using INIF, HEVC, and SIREN techniques, the overall filamentous shape can preserved, but with very different performance in preserving the striped patterns. For example, SIREN tends to produce excessively smoothed images that lack these finer details, while the HEVC approach introduces a significant amount of noise into the output. In contrast, INIF effectively retains the striped patterns to permit further sarcomeres analysis~\cite{craig2004molecular}. This can be further confirmed by comparing the segmentation on the decompressed images, where INIF achieved balanced results with minimal false positives and false negative errors.

\begin{figure}[htb]
    \centering
    \includegraphics[width=\textwidth]{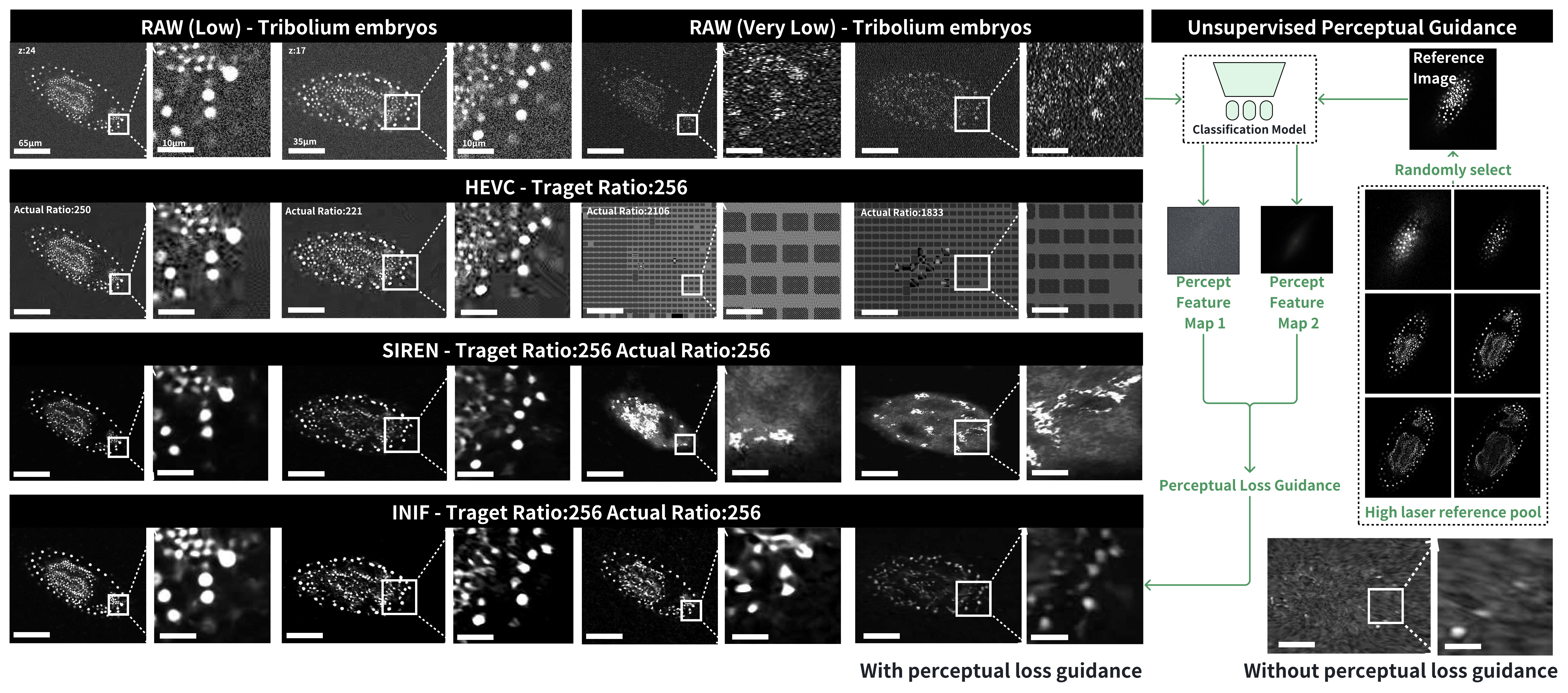}
    \caption{\textbf{INIF with perceptual guidance on noisy red flour beetle embryo data.} HEVC exhibits severe block artifacts, particularly on structures with lower contrast (left rows 2), while SIREN and INIF perform well in both high-contrast and low-contrast structures (left rows 3, 4). When dealing with extremely noisy data captured using very low laser power settings, INIF with perceptual-level guidance (row 4 middle) achieves better robustness compared to HEVC, SIREN, and INIF without perceptual guidance (right).}
    \label{fig6}
\end{figure}

From the results above, the INR-based compression method can mitigate high-frequency noise within the data to a certain extent, compared to classic codecs, like HEVC. This phenomenon can be attributed to the inherent characteristic of neural networks, which tend to learn continuous functions~\cite{xie2022neural}, particularly in compression tasks where model size is constrained. In this context, noise represents a challenging target for modelling purposes~\cite{al1998lossy}. Therefore, we want to demonstrate the flexibility of INIF to simultaneously compress the data and reduce the noise, i.e., only the important signals are compressed. This would find great potential in biological microscopy images, where effective image restoration is crucial in many applications, like live cell imaging with low laser power. To this end, INIF can be coupled with an additional perceptual loss function. Specifically, during compression, we incorporated additional reference data from relatively clear samples into each training iteration by selecting two random areas: one from the reconstruction by the INR network and another from the clear reference data. These two areas were then passed through the same classification network~\cite{zhang2018unreasonable} to obtain distinct perception feature maps. Subsequently, we calculated the perceptual loss between two feature maps as part of a joint optimization process aiming to simultaneously improve pixel-level similarities to noisy raw data while considering perception-level similarities to clear reference data. To verify the performance, we tested on confocal microscopy recordings of developing \textit{Tribolium casta-neum} (red flour beetle) embryos with three levels of laser power: high, low, and very low~\cite{weigert2018content}. Specifically, we conducted tests on samples captured at low and very low laser power settings (\ref{fig6}). As shown in Figure \ref{fig6}, the results indicate that when faced with noticeable noise at low laser power, HEVC exhibited severe block artifacts, particularly on structures with lower contrast. Without perceptual guidance, INIF could suffer from similar information loss as in SIREN. The perceptual loss encourages INIF to focus on the most critical information and therefore ignore the noise in the process of compression.

\subsection{Faster INIF with CODEC priors}\label{sec2.3}

\begin{figure}[htb]
    \centering
    \includegraphics[width=0.9\textwidth]{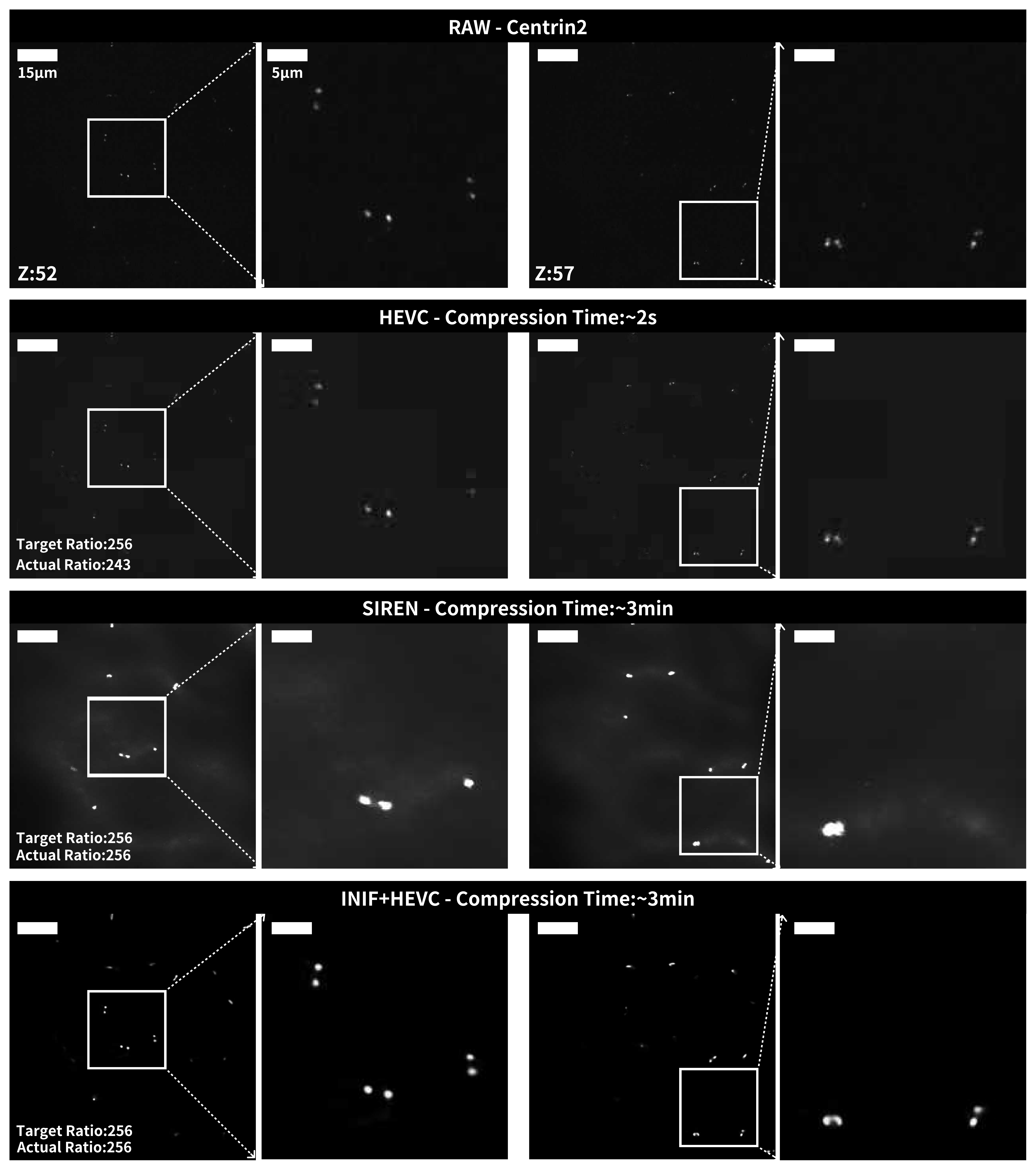}
    \caption{\textbf{INIF with CODEC prior on CENT2-stained data.} HEVC compression results in noticeable block artifacts, obscuring critical dot-like details (row 2). SIREN-compressed images suffer from incomplete convergence, leading to significant blurring (row 3). INIF leverages CODEC's prior knowledge to expedite training and achieves clearer images compared to HEVC even with a short training period (row 4).}
    \label{fig7}
\end{figure}

Despite the limitations of the codec in achieving optimal compression results in bioimaging, its significant advantage of minimal compression costs, both in terms of time and computation~\cite{sze2014high}, compels us to explore the possibility of incorporating it into the INIF framework to enhance compression efficiency. In most cases mentioned above, the primary advantage of the INIF framework has been its versatility and effectiveness in compression. However, iterative training unavoidably slows down the compression process. To address this issue, INIF has a hybrid mode to incorporate classic codec when fast compression is important. Specifically, we replace the raw image ground truth with residuals obtained by subtracting CODEC-decompressed images from raw images. Here, the INR network is used as an adapter~\cite{xu2023parameter} for CODEC's output. We not only enable adjustable compression objectivity to reduce blocky artifacts associated with CODEC but also accelerate the training process of INIF by reducing the amount of information to learn. 

We demonstrated the advantages of the hybrid mode of INIF by testing on 3D 100x confocal microscopy images of the Centrin-2 (CENT2) protein marking the centrosome in hiPS cells~\cite{viana2023integrated}. The most distinctive morphological feature of CETN2 in the image is isolated or pairs of bright blobs that are specifically localized to the centrioles varying at different cell cycle stages. Our analysis revealed that HEVC compression obscured critical puncta. Within a comparable amount of compression time,
SIREN-compressed images suffered from incomplete convergence of the loss function due to limited training epochs, leading to significant blurring that hindered practical application. However, INIF leveraged CODEC priors to expedite compression and achieved clearer images even with a short training period (Figure~\ref{fig7}).

\section{Discussion}\label{sec3}
The INIF workflow presents an alternative approach to CODEC for compressing and storing biological microscopy images. A key contribution of INIF is the integration of learned optimizers, a powerful INR training pipeline, and customizable application-appropriate guidance. With INIF, large microscopy data can be compressed with a controllable high compression ratio, such as 256-fold, and stored as an INIF file that supports arbitrary pixel-wise decoding. It is critical to preserve crucial information within bioimages after compression. Our INIF pushes the boundaries of microscopy image compression by offering larger compression ratios, enhanced compression quality, and adjustable objectivity throughout the differentiable compression process. We conduct various experiments to support this finding. With INIF, not only information about larger structures such as nuclei and cell membranes can be preserved after compression but also small structures can be identified and retained. Moreover, by incorporating application-appropriate guidance such as perceptual loss and segmentation loss, INIF achieves greater robustness on noisy data and assigned downstream tasks. Finally, we provide a hybrid mode by utilizing INIF as a final step adapter for any compressor; this makes it a versatile tool capable of expanding into broader applications.

However, INIF is still a lossy compressor, and cannot retain every bit of information. For instance, when compressing breast tumour tissue data, although we achieve better visualization results compared to HEVC and SIREN, it remains challenging to preserve all informative signals, particularly at higher compression ratios. Therefore, we strongly recommend using lossless compression methods~\cite{gupta2017modern} if the compression performance cannot meet the requirement for the downstream analysis. This leads us to our final discussion on ensuring the reliability of decompression quality. Fortunately, unlike generative tasks such as image restoration~\cite{weigert2018content} and super-resolution~\cite{chen2021three}, compression always has access to the raw data ground truth (e.g., saving on low-cost cold storage for backup purposes). Nevertheless, instead of simply comparing metrics like PSNR and SSIM~\cite{hore2010image} in a naive manner, we have demonstrated that an effective evaluation approach needs to focus on the overall biology application. For example, in the TNNI1 segmentation task, our assessment encompasses not only the visualization quality of decompressed images but also concerns regarding similarity with segmentation results (Figure \ref{fig5}). In the case of red flour beetle data, despite INIF exhibiting lower PSNR and SSIM values compared to HEVC in both low and very low cases; it becomes evident that INIF yields superior results for application purposes (Figure \ref{fig6}).

\section{Conclusion}\label{sec4}
To summarize, the primary objective of INIF is to raise awareness of efficient bioimaging data storage and provide a practical solution within existing hardware infrastructure. We investigated the major limitations of commercially available CODEC compressors when dealing with bioimages. Based on these shortcomings, we developed INIF as a compressor that supports compression and decompression of microscopy data with arbitrary shapes, allowing adjustable objectives for each compression task. We believe these functionalities guide future studies on bioimaging data compression to prioritize the adaptability of the compressor and the reliability of the decompressed data. This fundamental difference distinguishes scientific research-oriented data compression from industry-oriented applications.

\bibliography{sn-bibliography}

\section{Method}\label{sec5}
For each of the described experiments and case studies, we (1) extracted suitable data from private or public datasets, (2) trained an implicit neural representation (INR) network using a learned optimizer with or without additional guidance based on the desired functionality to be demonstrated, and (3) employed the trained INR network for decoding to decompress the data, subsequently quantifying and reporting the obtained results. Please refer to Supplementary \ref{Supp: exp} for additional details.

\subsection{Architectures of the INR network and learned optimizer}\label{sec5-1}
The SIREN network is a simple multi-layer perceptron (MLP) with a cosine activation function. For all our experiments, we utilized a 7-layer SIREN~\cite{sitzmann2020implicit} as the backbone of our Implicit Neural Representation (INR) network. We made some minor adjustments, such as introducing a learnable cosine frequency and refining the initialization strategy. One of our key innovations lies in the adoption of learned optimizers, which leverage artificial neural networks (ANNs) to update ANNs instead of relying solely on hand-crafted algorithms. In this study, we employed the versatile learned optimizer VeLO~\cite{metz2022velo}, which has been pre-trained at scale for optimization tasks and we further fine-tuned specifically for compression tasks. Our algorithm was implemented using the Google JAX framework~\cite{jax2018github}. Please refer to Figure~\ref{supfig1} for detailed model configurations.

\subsection{Data prepossessing}
 The initial step involves normalizing the input images to ensure consistency in their range. We have implemented Min-Max normalization~\cite{patro2015normalization}, which entails defining the minimum and maximum intensity values and subsequently mapping all values linearly within this range. Specifically, we have set the intensity range from 0 to 100. Before training, another preprocessing step is performed wherein we derive coordinates by extracting shape metadata from the target data and initializing coordinates with identical shapes corresponding to the data. Furthermore, these coordinates are normalized within each dimension to a range of -1 to 1. 

\subsection{Segmentation loss}
We utilize the Allen Cell Structure Segmenter (Segmenter) ~\cite{chen2018allen} to extract the segmentation map from images. The workflow of Segmenter involves several steps for image preprocessing and segmentation, including intensity normalization, edge-preserving smoothing, 3D filament filtering, and removal of small objects. Parameters for contrast enhancement were set as [1.5, 10.5], edge-preserving smoothing intensity was set to 1, F3 Params for filament detection were set as [1, 0.01], and remove small object intensity was set to 20. The segmentation loss is calculated using the intersection over union (IoU) Loss between the segmentation results of original and decompressed images. IoU can be formulated as follows:
\begin{equation}
    \text{IoU} = \frac{\text{Intersection}}{\text{Union}} = \frac{ M_{orignal}\cap  M_{decompressed} }{ M_{orignal}\cup  M_{decompressed} }
\end{equation}

\subsection{Perceptual loss}
The classification model chosen for extracting perceptual feature maps is the learned perceptual image patch similarity (LPIPS) ~\cite{zhang2018unreasonable}, which quantifies the discrepancy between the reference and reconstructed data. Specifically, we utilize the Alexnet version of LPIPS, which has been pre-trained on a natural image classification task. To further enhance its performance on bioimaging data, we fine-tune the model using a small amount of slice data from the hiPSC dataset. 

During the noisy data compression task, we randomly select a batch of pixel values from the clear reference and reconstruct an equal number of pixel values using the INR network. These two random image patches are then passed through the fine-tuned LPIPS model. The perceptual loss is defined as follows:

\begin{equation}    
\mathcal{L}_{\text{p}} = \sum_{i=1}^{L} \frac{1}{H_iW_i} \sum_{H_i,W_i}^{} \| w_i \cdot (\Phi_i(\hat{y}) - \Phi_i(y)) \|_2^2
\end{equation}

In this context, $\mathcal{L}_{\text{p}}$ represents the perceptual loss, $\hat{y}$ refers to the reconstructed patch, $y$ represents the reference patch, $\Phi_i(\cdot)$ denotes the feature representation at layer $i$, and $w_i$ is used as a scale factor to normalize each feature map in fine-tuned LPIPS model.

\subsection{CODEC prior}
The high-efficiency video coding (HEVC) standard~\cite{sze2014high} is specifically designed to provide an efficient encoding solution for transmitting high-definition video content. However, the absence of differentiable operators in HEVC impedes its adaptability to biological microscopy. Our objective is to leverage the cost-effective CODEC priors to enhance the efficiency of our methodology. Specifically, we allocate 90\% of the bitstream for preliminary compression using HEVC, followed by computing the residual image through differencing between the original image and decompressed output from HEVC. The remaining 10\% of the bitstream is utilized to initialize INIF, which is then trained to compress this residual image. Through this process, we effectively leverage the differentiable nature of INIF to facilitate adaptive learning objectives in CODEC, along with employing a filtered residual target to ease the training of the INR network. Please refer to Supplementary \ref{Supp: codec} for more detail.

\subsection{3D DNA stained hiPSC data}
Human induced pluripotent stem cells (hiPSCs) are a type of stem cell that has been artificially generated in the laboratory through the reprogramming of adult cells to attain an embryonic stem cell-like state. The WTC-11 hiPSC Single-Cell Image Dataset v1 encompasses 25 distinct intracellular structures observed in hiPSCs, with each sample including a channel specifically stained for DNA~\cite{viana2023integrated}. We randomly selected 25 samples exhibiting diverse structures and compressed the DNA-stained channel.

\begin{table}[h]
\centering
\begin{tabular}{cc}
\hline
FOV ID & Demonstrated Z value\\
\hline
10728 & 27\\
2299 & 35\\
142674 & 28\\
6030 & 34\\
10684 & 26\\
13309 & 28\\
\hline
\end{tabular}
\caption{DNA-stained hiPSC data used in our experiment, shown in (\ref{fig1}).}
\label{tab:hipsc_structures}
\end{table}


\subsection{Multi-channel breast tumour data}
We utilized a 3-dimensional (X, Y, C) multi-channel fluorescence microscopy dataset of breast tumour samples obtained from the 1-TNBC dataset within the Image Data Resource (IDR) public dataset project ID \verb|2801| ~\cite{friedman2020cancer}. The 1-TNBC dataset comprises 200 breast tumour tissue samples. Each sample represents a multi-dimensional image with the following characteristics: (i) spatial dimensions of 4506 $\times$ 4506 pixels depicting the two-dimensional tissue structure, (ii) five channels labelling distinct cellular structures and biomarkers in the tumour samples, namely S100A4-Opal 520 image (labelling S100A4 protein), CK-Opal 570 image (labelling cytokeratins), PDPN-Opal 650 image (labelling podoplanin), DAPI image (labelling cell nuclei), and Bright-field image. In our experiment, we utilized \verb|Patient_55_core_1.tiff| file in (\ref{fig3}). 

\subsection{5D mouse organoid data}

This is a 5-dimensional (X, Y, Z, C, T) multi-channel fluorescence microscopy dataset that captures the growth dynamics of mouse organoids over time. It was obtained from the Zenodo repository ~\cite{Smith_Sparks_Almagro_Behrens_Dunsby_Salbreux_Chaigne_2023}. The dataset consists of 6 tiles (\verb|Tile_1| through \verb|Tile_6|) containing time-lapse 4D image sequences. Each tile represents a multi-dimensional image with specific properties: (i) spatial dimensions of 135 $\times$ 136 $\times$ 160 representing the volumetric structure in three dimensions; (ii) two fluorescence channels labelling DNA (red channel) and cell membrane (green channel); and (iii) 135 timepoints capturing the temporal evolution of organoid morphogenesis. For visualization and validation purposes, we utilized a subset of the \verb|Tile_1_processed_binned-2b| file by extracting twelve sub-volumes as shown in the table below:

\begin{table}[h]
\centering
\begin{tabular}{ccc}
\hline
Demonstrated C value & Demonstrated T value & Demonstrated Z value\\
\hline
1 & 32 & 60\\
1 & 32 & 80\\
1 & 32 & 120\\
2 & 32 & 60\\
2 & 32 & 80\\
2 & 32 & 120\\
1 & 97 & 60\\
1 & 97 & 80\\ 
1 & 97 & 120\\
2 & 97 & 60\\
2 & 97 & 80\\
2 & 97 & 120\\
\hline
\end{tabular}
\caption{The visualization sub-volumes extracted from the 5D mouse organoid dataset for validation are presented in (\ref{fig3}), with indexing based on channel, timepoint, and z-plane.}
\label{tab:mouse_organoid_subvols}
\end{table}


\subsection{Noisy Tribolium data}
This dataset comprises confocal microscopy recordings of developing \textit{Tribolium castaneum} (red flour beetle) embryos, obtained from ~\cite{weigert2018content}, with three levels of laser power: high, low, and very low. All images were acquired using the Zeiss 710 multiphoton laser scanning microscope equipped with a 25x multi-immersion objective. In our experiment, we evaluate the performance of our proposed method on two samples captured at low and very-low laser power from this dataset.
\begin{table}[h]
\centering
\begin{tabular}{lccc}
\hline
File Name & Laser Power & Dimensions \\
\hline
\verb|nGFP_0.1_0.2_0.5_20_13_late|& Low & {665 x 773 x 48}  \\
\verb|nGFP_0.1_0.2_0.5_20_14_late| & Low & {486 x 954 x 45}  \\
\verb|nGFP_0.1_0.2_0.5_20_13_late| & Very Low & {665 x 773 x 48}  \\
\verb|nGFP_0.1_0.2_0.5_20_14_late| & Very Low & {486 x 954 x 45}  \\
\hline
\end{tabular}
\caption{Subset of images from the noisy Tribolium dataset used in our experiment, shown in (\ref{fig6})}.
\label{tab:tribolium_dataset}
\end{table}

\subsection{Centrin-2 stained hiPSC 3D data}
In this experiment, we utilized the structure-stained data of Centrin-2 from the WTC-11 hiPSC Single-Cell Image Dataset v1 to investigate its association with the centrosome. The centrosome, composed of two centrioles surrounded by pericentriolar material (PCM), plays a crucial role in cell division by duplicating and aiding in the formation of the mitotic spindle for proper chromosome segregation.

\subsection{TNNI1 stained hiPSC 3D data}


This data is a 4-dimensional (X, Y, Z, C) multi-channel fluorescence microscopy data that comprises the three-dimensional structural data of the Troponin I Type 1 (TNNI1) protein obtained from fluorescently tagged human induced pluripotent stem cell (hiPSC) lines at the Allen Institute for Cell Science ~\cite{chen2018allen}. This dataset represents a multi-dimensional image with specific properties: (i) spatial dimensions of 555 × 441 × 60 representing the volumetric structure in three dimensions; (ii) four fluorescence channels: DNA channel, cell membrane channel, TNNI1 protein structure channel, and bright field channel. For visualization and validation purposes, we utilized the TNNI1 protein structure at depth 30.

\section{Code Availability}

Code to reproduce the experiements in this work or to compress your own data can be found at \url{https://github.com/PKU-HMI/INIF}.

\newpage
\begin{appendices}

\section{Supplementary Tables}
\subsection{Supplementary Table 1}

\begin{table}[h]
\label{Sup: Table1}
\caption{Overview of compared compression methods and their implementations}
\begin{tabular}{llll}
\hline
Method & GPU & Language/Source \\ \hline
HEVC & NO & C/C++ / \url{https://www.videolan.org/developers/x265.html} \\
SIREN & YES & Python / \url{https://github.com/lucidrains/siren-pytorch} \\
INIF (ours) & YES & Python / \url{https://github.com/PKU-HMI/INIF} \\ \hline
\end{tabular}
\end{table}

\subsection{Supplementary Table 2}

\begin{table}[h]
\caption{Overview of case studies and experiments}
\label{Sup: Table2}
\begin{tabular}{p{1cm}@{\hspace{0.5cm}}p{2.5cm}@{\hspace{0.5cm}}p{2cm}@{\hspace{0.5cm}}p{3cm}@{\hspace{0.5cm}}p{3cm}}
\hline
Figure & Dataset & Task & Biological Structure & Microscope \\ \hline
2 & WTC-11 hiPSC Single-Cell Image & Compression & 25 cellular structures & Spinning-disk confocal microscope \\
3 & 1-TNBC & Compression & breast tumour & Multi-channel fluorescence microscopy \\
4 & Mouse Organoids & Compression & Organoids & Timelapse dual oblique plane microscopy \\
5 & Troponin I & Compression (segmentation guidance) & Type 1 TNNI1 & Spinning-disk confocal microscope \\
6 & Tribolium & Compression (denoising guidance) & Whole embryo of Tribolium castaneum & Confocal \\ \hline
\end{tabular}
\end{table}

\section{Supplementary Notes}
\subsection{Background}
\label{Sup: Background}
\subsubsection{Implicit Neural Compression }

\begin{figure}[htb]
    \centering
    \includegraphics[width=\textwidth]{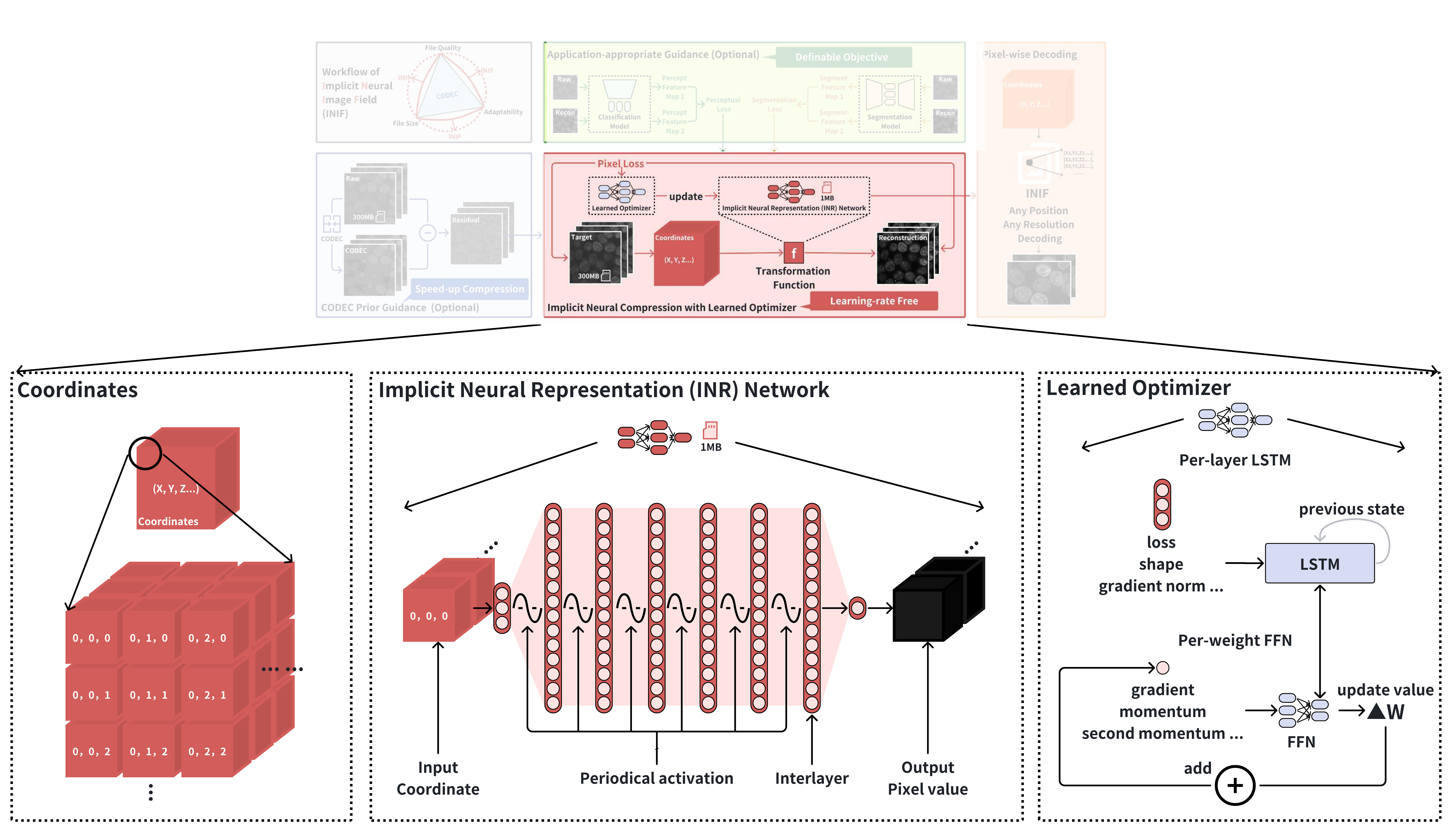}
    \caption{\textbf{Pipeline for INR-based Compression
.} Starts with creating the coordinates with as the same shape of the data, followed by iterative passing those coordinates into a small INR network. The output of INR network would be the predicted pixel values and are further compared with the raw data value for back-propagation calculation. Values are then sent to the learned optimizer to predict the updated value for each parameter. More detail on learned optimizer can be found in Section~\ref{sec:sup:Learned Optimizer}}
    \label{supfig1}
\end{figure}

Implicit Neural Representations (INRs), introduced by \cite{chen2021_learning_continuous_image_representation_with_local_implicit_image_function, zhao2023moec, sitzmann2020_siren_implicit_neural_representaions_with_periodic_activation_functions, mildenhall2021_nerf}, have emerged as a powerful method for data representing, such as images \cite{thies2019_deferred_neural_rendering_image_synthesis_using_neural_textures}, videos \cite{NEURIPS2021_b4418237, xian2021space}, and audios \cite{su2022inras, lanzendorfer2023siamese}. INRs leverage neural networks to approximate a compact, continuous function, enabling efficient representation of high-dimensional data. This ability has led to the development of implicit neural compression, which has shown promising results for efficient image compression by encoding images as compact neural network weights \cite{strumpler2022_inr_for_image_compression}. The INR network training pipeline of INIF is shown in Figure~\ref{supfig1}.

The core idea behind implicit neural compression is to store all information implicitly in the network weights $\theta$. During training, it trains a neural network $f_\theta$ parameterized by weights $\theta$, typically a coordinate-based multi-layer perceptron (MLP) \cite{popescu2009_multilayer}, to map low-dimensional coordinate inputs $\mathbf{x} \in \mathbb{R}^n$ (e.g., spatial coordinates for images or space-time coordinates for videos) to the corresponding high-dimensional signal values $\mathbf{y} \in \mathbb{R}^m$ (e.g., grayscale pixel values for bioimages or RGB-alpha values for videos with transparency):

\begin{equation}
f_\theta: \mathbb{R}^n \rightarrow \mathbb{R}^m, \quad \mathbf{y} = f_\theta(\mathbf{x})
\end{equation}

During training, the neural network learns to overfit the underlying data distribution by minimizing a reconstruction loss $\mathcal{L}$ between the network output and the target data:

\begin{equation}
\min_\theta \mathcal{L}(f_\theta(\mathbf{x}), \mathbf{y}^\text{target})
\end{equation}

Normally, the reconstruction loss $\mathcal{L}$ is chosen to use Mean Squared Error(MSE) to measure the similarity of the ground-truth target and the INRs output. Once the network is trained, we only need to store the network weights $\theta$ because they constitute a continuous and compressible representation of the data. To reconstruct the original data or decompress it one simply evaluates the trained function $f_\theta$ at the desired input coordinates $\mathbf{x}$, thereby obtaining the corresponding signal values $\mathbf{y}$. 

However, implicit neural compression also faces challenges. The optimization process to fit the network to the target data can be computationally expensive, especially for high-resolution images or videos. Additionally, the compression performance may vary depending on the data content and the architecture of the neural network used. To address these challenges, recent works have explored various techniques to improve the efficiency and performance of implicit neural compression. These techniques include improving network structures \cite{chen2021_nerv_neural_representation_for_videos, fathony2020_multiplicative_filter_networks, lu2021_compressive_neural_representations_of_volumetric_scalar_fields}, selecting appropriate activation functions \cite{klocek2019_hypernetwork, mehta2021_modulated_periodic_activations, sitzmann2020_siren_implicit_neural_representaions_with_periodic_activation_functions}, and employing embeddings \cite{benbarka2022_seeing_inr_as_fourier_series, muller2022_instant_neural_graphics}.

\subsection{Learned Optimizer}\label{sec:sup:Learned Optimizer}
\subsubsection{Learned Optimization }

In traditional machine learning optimization methods, hand-designed optimizers are typically used to train networks. A significant issue with hand-designed optimizers is the need for meticulous tuning of hyperparameters across different tasks. Without such tuning, training can fail or become trapped in local optima. The learned optimizer is a novel approach that attempts to use neural networks to predict the updates for the network being trained. The concept of meta-learning update rule can be traced back to the work of Bengio et al. \cite{Succ2002On}and Runarsson and Jonsson \cite{runarsson2000evolution}, which both trained simple neural networks to learn simple update rule. 

In 2016, Andrychowicz et al. \cite{andrychowicz2016learning} meta-trained an RNN-parameterized learned optimizer on deep learning tasks by backpropagating through the optimization process, and this work sparked a significant wave of research, leading to the development of various new techniques.
Wichrowska et al. \cite{wichrowska2017learned} introduced hierarchical learned optimizers. Their approach involved meta-training these optimizers on a broad range of synthetic tasks. More recently, Metz et al. \cite{metz2020tasks} extended this line of research by focusing on a more realistic task distribution and enhancing the learned optimizer architecture \cite{metz2020using}. Finally, in our work, we used the Versatile Learned Optimizers (VeLO) provided by Metz et al. \cite{metz2022velo}, which was meta-trained at a far greater scale than the previous investigation.

\subsubsection{VeLO}

We can define the optimization problem in the following form ~\ref{sup_b_eq1}:
\begin{equation}
    \phi_{t+1} = \phi_{t} - U(\phi_{t}, \nabla_{\phi_{t}}; \theta)
    \label{sup_b_eq1}
\end{equation}
where $\phi$ represents the parameters of the INR being optimized, $\nabla$ denotes the gradient of the loss function of the INR network obtained through backpropagation, $\theta$ are the parameters of VeLO, and $U$ is the update predicted by VeLO.

As shown in Figure ~\ref{supfig1}, the core module of VeLO is hierarchical and consists of two components: a per-tensor LSTM and a per-parameter MLP.

The per-tensor LSTM is composed of 512 hidden units and is responsible for managing the overall training features. It serves as a hypernetwork for the per-parameter MLP by providing individualized parameters for the per-parameter MLP. Its input features reflect the overall training state and the statistical characteristics of each parameter, including the following features: (1) Fraction of training remaining: This feature uses the current iteration $t$ and he total target iterations $T$ to compute a set of training soft progress values using the formula ~\ref{sup_b_eq2}:
\begin{equation}
    \label{sup_b_eq2}
    \tanh(10 \times (t/T - s))
\end{equation}
(2) Loss features: These features use the exponential moving average and running minimum of the loss to construct features that reflect the training trend, independent of the loss magnitude. The computed values range from $[-1, 1]$, where negative values roughly correspond to decreasing loss, positive values indicate increasing loss, and zero indicates no change in loss, (3) first and second moment of momentum features, and (4) the rank of the tensor. Finally, the per-tensor LSTM outputs value $d$ and scalar learning rate $c_{lr}$, which are used to update the per-parameter MLP as follows equation ~\ref{sup_b_eq3}:
\begin{equation}
    \label{sup_b_eq3}
    \Delta\theta_{p} = 0.001 \times d \times \exp(0.001 \times c_{lr})\|\theta_{p}\|_2
\end{equation}

The per-parameter MLP operates specifically on feature sets, using a minimal MLP (2-hidden layers, 4-hidden units), which remains computational efficiency with high performance. For all parameters in the INR we want to optimize, VeLO initializes a unique hyper MLP (hMLP). The initialization values are determined by a uniform distribution and meta learned parameter weights. The hMLP predicts an update $U_\theta$ after receiving the gradient, momentum, second moment, and other features of the INR network, in order to update the INR network weight $\phi$ in formula ~\ref{sup_b_eq1}.

\subsection{Codecs}
\label{Supp: codec}
\subsubsection{Hybrid coding framework}

\begin{figure}[htb]
    \centering
    \includegraphics[width=\textwidth]{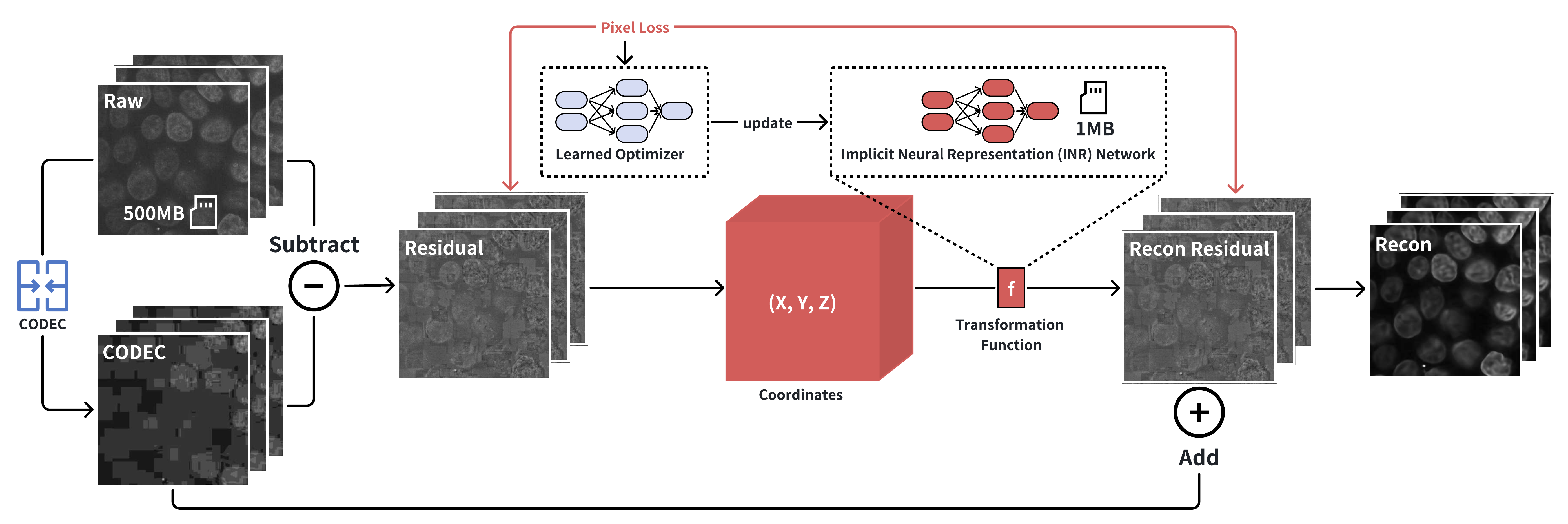}
    \caption{\textbf{Codec prior for speed-up INR compression.} Codec can produce a reasonable result with a relatively cheap cost. Consider this as a start-point, INR networks are utilized as the differentiable adapter to further adapt the result.}
    \label{supfig2}
\end{figure}

\begin{figure}[htb]
    \centering
    \includegraphics[width=\textwidth]{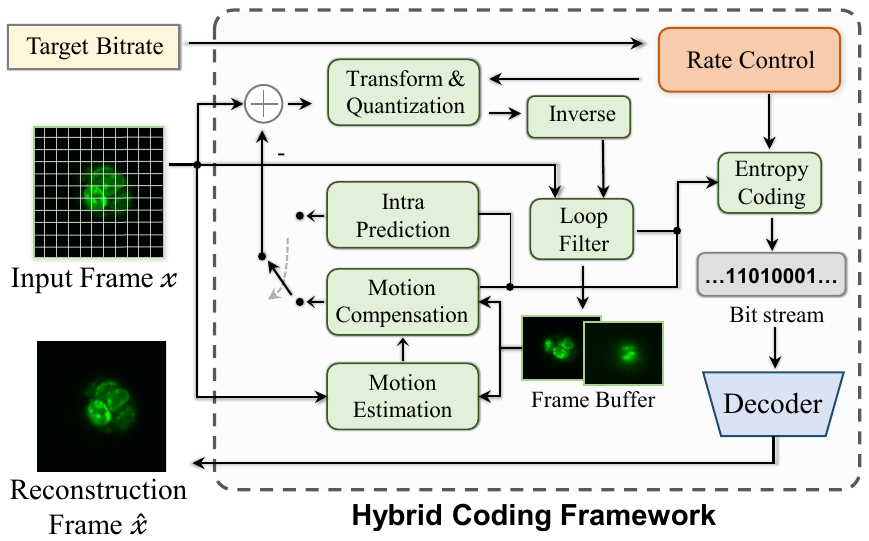}
    \caption{\textbf{HEVC Hybrid Coding Framework.} Inputting multi-frame data, the HEVC encoder performs intra-frame and inter-frame prediction respectively to eliminate redundant information within and between images.}
    \label{supfig3}
\end{figure}

One of our core designs shown in Section~\ref{sec2.3} uses codec priors as a solid starting point before INR compression (Figure~\ref{supfig2}). Current commercial coding standards for image and video compression, such as H.264/AVC \cite{wiegand2003overview} and H.265/HEVC~\cite{sullivan2012overview}, are based on hybridizing a broad range of algorithms for predictive coding. Those algorithms assume the presence of spatial and temporal redundancy in pictures and videos. During the compression process, the codec uses pixels in adjacent areas to predict the current area and then records the residual between the true value and the predicted value to achieve data compression. As shown in Fig.~\ref{supfig3}, the HEVC standard follows the hybrid coding framework. The encoding steps of HEVC are mainly divided into block division, intra-prediction, inter-prediction, transformation, quantization, post-processing filtering, and entropy coding. 

During the execution of encoding, the HEVC codec considers the results of different encoding tools and encoding parameters:
\begin{equation}
    \mathop{\arg\min}\limits_{\theta} (D + \lambda \cdot R) , \label{eq_rdo}
\end{equation}
where $D$ represents the coding distortion, $R$ represents the bitrate required to encode the information, and $\lambda$ represents the Lagrange multiplier to balance $D$ and $R$. $\theta$ represents the combination of coding tools and coding parameters. Codec aims to minimize Eq.~\ref{eq_rdo}. This process is called Rate-distortion Optimization~(RDO)~\cite{gao2017temporally,cassa2012fast} and is the basic principle of codec encoding.

For each frame of input data, the HEVC codec first divides it into Coding Tree Units~(CTUs) according to the predefined fixed size. Then, the codec further divides the units according to the complexity of the image texture to obtain Coding Units~(CUs), Prediction Units~(PUs), and Transform Units~(TUs) for subsequent operations. Considering the existence of spatial redundancy, which means the pixel values of adjacent areas are similar, the HEVC codec can predict the pixel value of the current PU with reference to the encoded intra-pixels. When pictures are taken continuously, there are also similarities between pixels at adjacent positions in adjacent frames. Therefore, the HEVC codec can also use units of adjacent frames to perform inter-frame prediction on the current PU. The residual between the predicted value and the true value of the unit will be transformed to the frequency domain by codec, and quantization with different step sizes will be performed according to different frequency bands. This reduces the amount of data that needs to be stored while minimizing perceptible distortion. Quantization introduces distortion to the image. Therefore, the codec performs in-loop filtering on the reconstructed image to alleviate artificial artifacts such as blocking and ringing introduced during the lossy compression process. Finally, all the above coding modes and data information are written into the bit stream in a lossless manner by the entropy coding module.

\subsubsection{Rate Control}
As an important tool for codec control, the rate control algorithm can make the bitrate of the encoder output stream consistent with the specified target, thereby realizing the function of controlling the compression rate~\cite{li2014lambda}. Before encoding, the target bitrate for rate control can be obtained based on the original size of the data and the target compression rate. Given a target bitrate, the rate control algorithm maps it to encoding parameters for different frames and CTUs, such as Quantization Parameter~(QP) and $\lambda$ in Eq.~\ref{eq_rdo}. QP achieves different distortions by controlling the quantization step size of the unit. $\lambda$ achieves different compression rates by adjusting the weight of the unit's distortion and bitrate during the RDO process.

\begin{figure}[htb]
    \centering
    \includegraphics[width=0.5\textwidth]{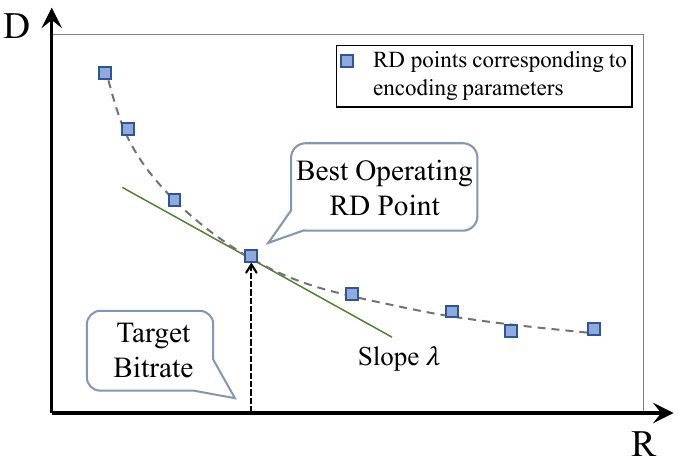}
    \caption{\textbf{Rate-distortion Curve of Coding Unit.}}
    \label{supfig4}
\end{figure}

Basically, rate control is divided into bit allocation and coding parameter derivation. Based on the target bitrate, frame size, and frame rate, the average number of bits consumed per frame can be calculated. Taking further into account the importance of different frames in the reference process and the actual consumption and surplus of bits by encoded frames, bits are adaptively allocated to frames and CTUs. For the coding unit, the bitrate and distortion have an inverse relationship as shown in Fig.~\ref{supfig4}. In HEVC, the relationship between $R$ and $D$ is modeled as a hyperbolic function:
\begin{equation}
    D(R) = C \cdot R^{-K} , \label{eq_rd}
\end{equation}
where $C$ and $K$ are model parameters, which vary with coding unit content characteristics. Combining Eq.~\ref{eq_rdo} and Eq.~\ref{eq_rd}:
\begin{equation}
    \lambda = \frac{\partial D}{\partial R} = C \cdot K \cdot R^{-K-1} \triangleq \alpha \cdot R^{\beta} , \label{eq_lambda}
\end{equation}
where $\alpha$ and $\beta$ are model parameters that are dynamically updated to reflect the characteristics of the unit and the outcomes of the encoding process throughout the encoding period. With the help of Eq.~\ref{eq_lambda}, the bits are mapped to the encoding parameter $\lambda$ to adjust the RDO. QP can also be further derived:
\begin{equation}
    QP = c_1 \cdot \ln(\lambda) + c_2 , \label{eq_qp}
\end{equation}
where $c_1$ and $c_2$ are respectively set to 4.2005 and 13.7122 in HEVC.

\subsection{Compression Experiments}
\label{Supp: exp}
\subsubsection{Overview}
For each of the described experiments, our method consists of the following steps:
\begin{enumerate}
    \item \textbf{Image Normalization}: Prior to training, each image is normalized to ensure a consistent scale across different images. This step is critical as it helps to standardize the input data, improving the stability and performance of the training process.
    \item \textbf{Training for Compression}: For every selected image, an INR model is individually trained to compress that image. This process involves optimizing the INR to overfit the image directly into the model's parameters.
    \item \textbf{Reconstruction}: The trained INR models are then used to reconstruct the original images by feeding their coordinates.
In particular, each INIF model is trained on a specific image, capturing its unique content (e.g., nuclei, microtubules). We also used the auto-adjustment function in ImageJ for all of our results (including baselines) to better demonstrate the visualization qualities. 
\end{enumerate}

\subsubsection{Image normalization}
For both the training and prediction phases, normalizing input images to a standard range is crucial. This is because the compressed image $\mathbf{y}$ predicted by our network and the corresponding ground-truth image $\mathbf{y^{\text{truth}}}$ typically differ significantly in the dynamic range of their pixel values. Normalization is commonly achieved by scaling the inputs to a specified minimum and maximum value. In our work, we employ min-max normalization \cite{patro2015normalization} to scale each pixel value and coordinate within the specified range. The normalized image $N(\mathbf{y})$ and coordinate $N(\mathbf{x})$ are formulated as follows:
\begin{equation}
N(\mathbf{y}; \mathbf{y}_{\min}, \mathbf{y}_{\max}) = \frac{\mathbf{y} - \mathbf{y}_{\min}}{\mathbf{y}_{\max} - \mathbf{y}_{\min}}
\end{equation}

\begin{equation}
N(\mathbf{x}; \mathbf{x}_{\min}, \mathbf{x}_{\max}) = \frac{\mathbf{x} - \mathbf{x}_{\min}}{\mathbf{x}_{\max} - \mathbf{x}_{\min}}
\end{equation}

\noindent where pixel values are scaled to the range from 0 to 100, and coordinate values are scaled to range from -1 to 1.

\subsubsection{Image quality assessment}
Assessing the quality of image compression is critical to ensure that the compressed images maintain a high degree of fidelity to the original. To this end, we utilize several metrics to assess the quality of our image compression:
\begin{enumerate}
    \item \textbf{Pixel level loss}: We utilize Mean Squared Error (MSE) to quantify the average of the squares of the differences between the ground-truth image, $\mathbf{y}^{\text{truth}}$, and the compressed image, $\hat{\mathbf{y}}$. This metric is widely used in image processing to measure the quality of reconstructed images as it directly corresponds to pixel-wise intensity differences:
    \begin{equation}
       \operatorname{MSE}(\mathbf{y}^{\text{truth}}, \hat{\mathbf{y}}) = \frac{1}{n} \sum_{i=1}^{n} (\mathbf{y}^{\text{truth}}_i - \hat{\mathbf{y}}_i)^2,
    \end{equation}
    \noindent Here, $n$ represents the total number of pixels in the image.
    \item \textbf{The perceptual loss}: Perceptual loss ~\cite{johnson2016_perceptual_loss} measures the difference in perceptual features between two images, capturing aspects that traditional pixel-based metrics might miss. Initially, we used the AlexNet ~\cite{krizhevsky2012_alexnet_imagenet} model, pre-trained on the Allencell ~\cite{chen2018allen} dataset, to extract high-level semantic features from both the original and compressed images. This approach helps in assessing how well the compression model preserves content that is perceptually significant to human observers:
    \begin{equation}
        \mathcal{L}_p = \sum_{i=1}^L \frac{1}{H_i W_i} \sum_{h=1}^{H_i} \sum_{w=1}^{W_i} \|w_i \cdot (\Phi_i(\hat{\mathbf{y}}) - \Phi_i(\mathbf{y}^{\text{truth}}))\|_2^2,
    \end{equation}
    \begin{figure}[htb]
    \centering
    \includegraphics[width=\textwidth]{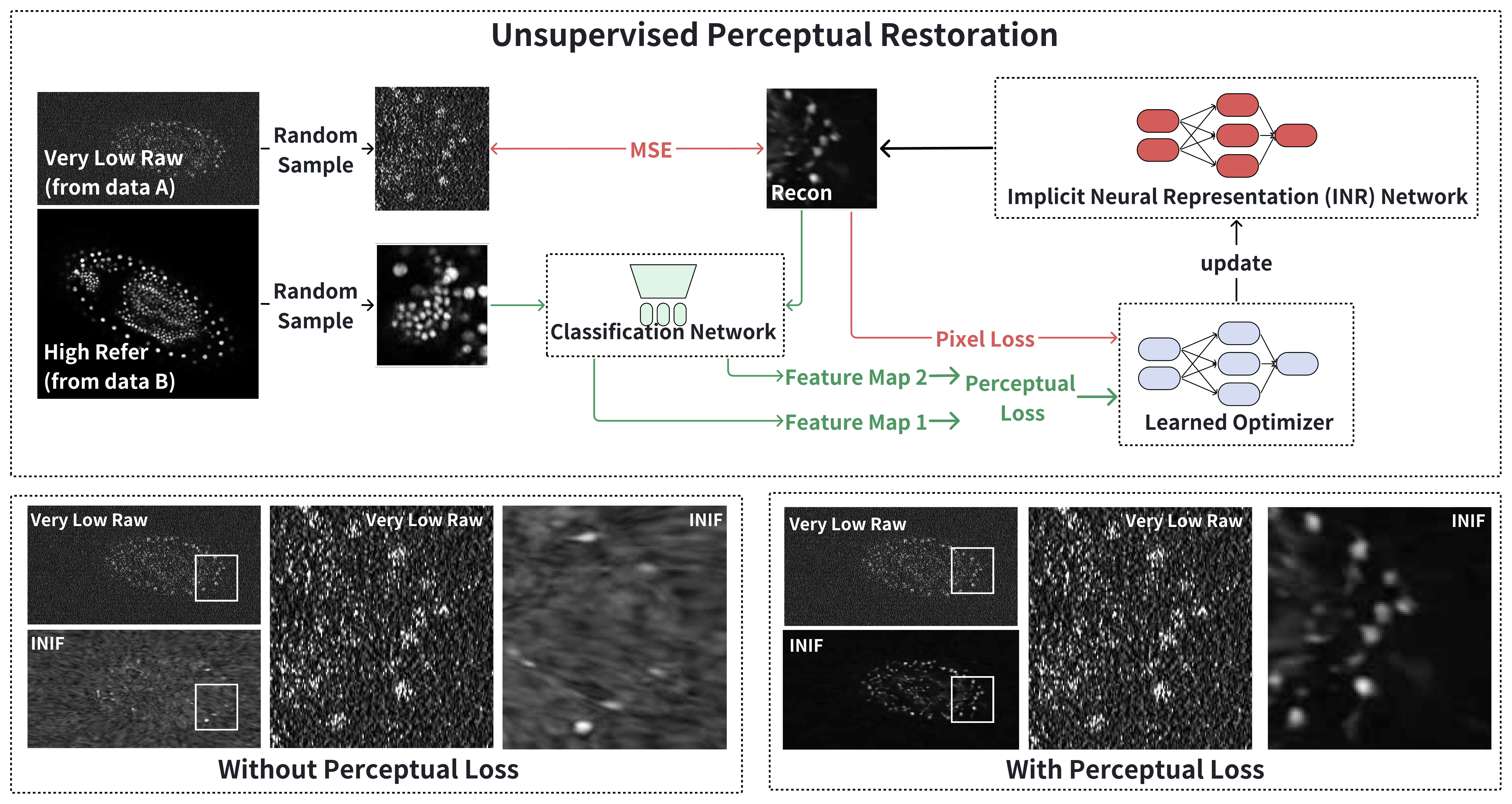}
    \caption{\textbf{Unsupervised perceptual restoration.}}
    \label{supfig5}
    \end{figure}
    where $\Phi_i$ denotes the feature map obtained from the $i$-th layer of AlexNet, and $w_i$ represents the weight factors for the respective layers. We use perceptual loss in our unsupervised robust noisy data compression experiment. Listed in Section~\ref{sec2.2}. The pipeline is shown in Figure~\ref{supfig5}  
    \item \textbf{The structural similarity index measure (SSIM)}: 
    SSIM \cite{wang2004image_ssim_image_quality_assessment} is an advanced metric that evaluates image quality by comparing the luminance $l(\mathbf{y}^{\text{truth}}, \hat{\mathbf{y}})$, contrast $c(\mathbf{y}^{\text{truth}}, \hat{\mathbf{y}})$, and structural $s(\mathbf{y}^{\text{truth}}, \hat{\mathbf{y}})$ similarities between the original $\mathbf{y}^{\text{truth}}$ and compressed images $\hat{\mathbf{y}}$. Unlike MSE, SSIM considers perceptual phenomena, making it more aligned with human visual perception:
    \begin{equation}
        \operatorname{SSIM}(\mathbf{y}^{\text{truth}}, \hat{\mathbf{y}}) = l(\mathbf{y}^{\text{truth}}, \hat{\mathbf{y}}) \cdot c(\mathbf{y}^{\text{truth}}, \hat{\mathbf{y}}) \cdot s(\mathbf{y}^{\text{truth}}, \hat{\mathbf{y}}),
    \end{equation}
    where:
    \begin{align}
    l(\mathbf{y}^{\text{truth}}, \hat{\mathbf{y}}) &= \frac{2 \mu_{\mathbf{y}^{\text{truth}}} \mu_{\hat{\mathbf{y}}} + c_1}{\mu_{\mathbf{y}^{\text{truth}}}^2 + \mu_{\hat{\mathbf{y}}}^2 + c_1}, \\
    c(\mathbf{y}^{\text{truth}}, \hat{\mathbf{y}}) &= \frac{2 \sigma_{\mathbf{y}^{\text{truth}}} \sigma_{\hat{\mathbf{y}}} + c_2}{\sigma_{\mathbf{y}^{\text{truth}}}^2 + \sigma_{\hat{\mathbf{y}}}^2 + c_2}, \\
    s(\mathbf{y}^{\text{truth}}, \hat{\mathbf{y}}) &= \frac{\sigma_{\mathbf{y}^{\text{truth}} \hat{\mathbf{y}}} + c_3}{\sigma_{\mathbf{y}^{\text{truth}}} \sigma_{\hat{\mathbf{y}}} + c_3}.
    \end{align}
    Here, $\mu_{\mathbf{y}^{\text{truth}}}$ and $\mu_{\hat{\mathbf{y}}}$ are the mean values of $\mathbf{y}^{\text{truth}}$ and $\hat{\mathbf{y}}$, $\sigma_{\mathbf{y}^{\text{truth}}}^2$ and $\sigma_{\hat{\mathbf{y}}}^2$ are their variances, and $\sigma_{\mathbf{y}^{\text{truth}} \hat{\mathbf{y}}}$ is the covariance. Constants $c_1$, $c_2$, and $c_3$ are small constants added to stabilize the division with a weak denominator.
\end{enumerate}

\section{Stress test on larger images}
We conducted stress tests on three large 3D TIFF images with the size of 4GB, 11GB, and 23GB, respectively. The 4GB image depicts a whole-mount mouse lung stained with CD3 (T cells) and CD19 (B cells), dehydrated with Ethanol, and cleared with ethyl cinnamate (ECi). The 11GB and 23GB images show the mouse gut injected with CD31 (vessels) and EpCAM (Epithelium) antibodies, then dehydrated with Ethanol and cleared with ECi.

\begin{figure}[htb]
    \centering
    \includegraphics[width=\textwidth]{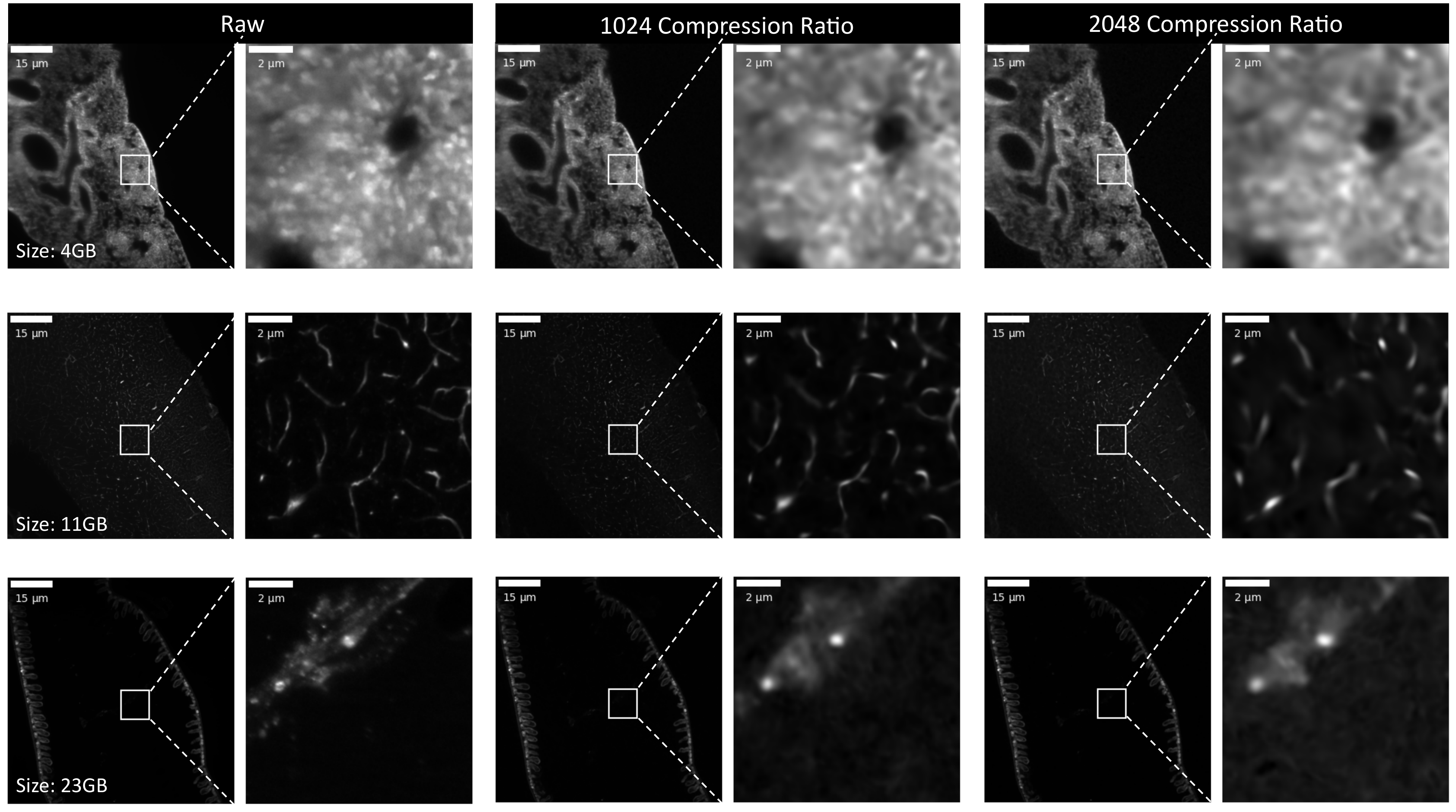}
    \caption{\textbf{INIF under large tiff images.} INIF using different compression ratios exhibits plausible compression ability among selected three large tiff images. }
    \label{fig_large_tiff}
\end{figure}

Due to memory constraints, we partitioned the large image into smaller sections, ensuring each section fits within the memory capacity (we used one 40GB NVIDIA A100 GPU). Chunking was performed along the z-axis. Consequently, the INIF algorithm was employed on each chunk utilizing two distinct compression ratios (1024, 2048), respectively. The results are shown in Fig \ref{fig_large_tiff}, which can verify the applicability.

\section{Decoding strategies of INIF}
\begin{figure}[htb]
    \centering
    \includegraphics[width=\textwidth]{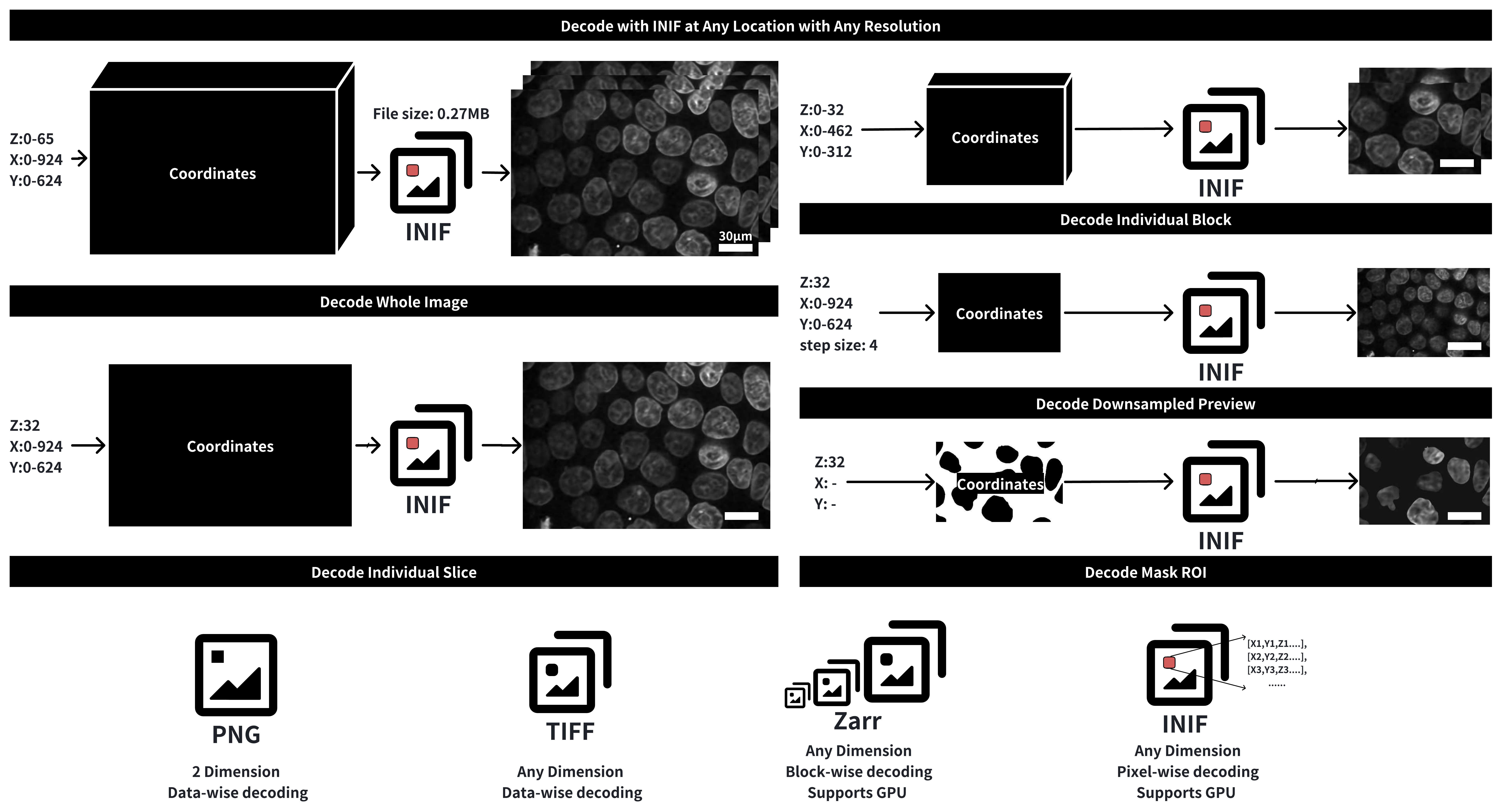}
    \caption{\textbf{Decoding INIF file in a pixel-wise fashion.} Unlike traditional file formats like TIFF that require complete decoding of the entire dataset, INIF enables pixel-wise decoding by providing the coordinates of the desired region of interest (ROI) to the INR network. INIF also supports multi-resolution decoding functionality and location-specific decoding even for irregularly shaped ROIs represented by binary masks.}
    \label{fig8}
\end{figure}

    Commonly used file formats (e.g., JPG and PNG) can only used for regular 2D RGB images. Specialized file formats such as Ome-TIFF or Zarr have been developed for storing and decoding high-dimensional microscopy images~\cite{abernathey2021cloud, hamzah2021jpg, wiggins2001image}. Comparing to Ome-Tiff, Zarr~\cite{abernathey2021cloud} offers more flexibility with additional features like multi-resolution storage, block-wise decoding, and parallel decoding using GPU. The evolution of these file formats indicates that the ultimate objective is to efficiently decode and visualize specific regions of interest (ROI)~\cite{chityala2004region} in any microscopy data as needed since full decoding can be time-consuming and memory-intensive. In this context, INIF provides an excellent solution by enabling pixel-wise decoding. Taking a 3D (XYZ) confocal microscopy images of DNA~\cite{viana2023integrated} as an example, assume our goal is to visualize slice 32. Decoding TIFF requires complete decoding of the entire dataset with dimensions X:924, Y:624, and Z:65 – including extracting all pixel values followed by selecting only those belonging to slice 32 for visualization purposes. On the other hand, leveraging INIF's pixel-wise decoding function allows us to simply provide the coordinates of slice 32 to the INR network for efficient decoding. Furthermore, by assigning a subset of coordinates, we can also support multi-resolution functionality – particularly useful when fast preview-based decodings are required without needing additional storage space for different resolution versions of the same data which is used in Zarr format. Additionally, even in special scenarios involving irregularly shaped ROIs represented by binary masks, we can still handle them through appropriate location-based decoding (\ref{fig8}).

\section{Limitations}
\begin{figure}[htb]
    \centering
    \includegraphics[width=\textwidth]{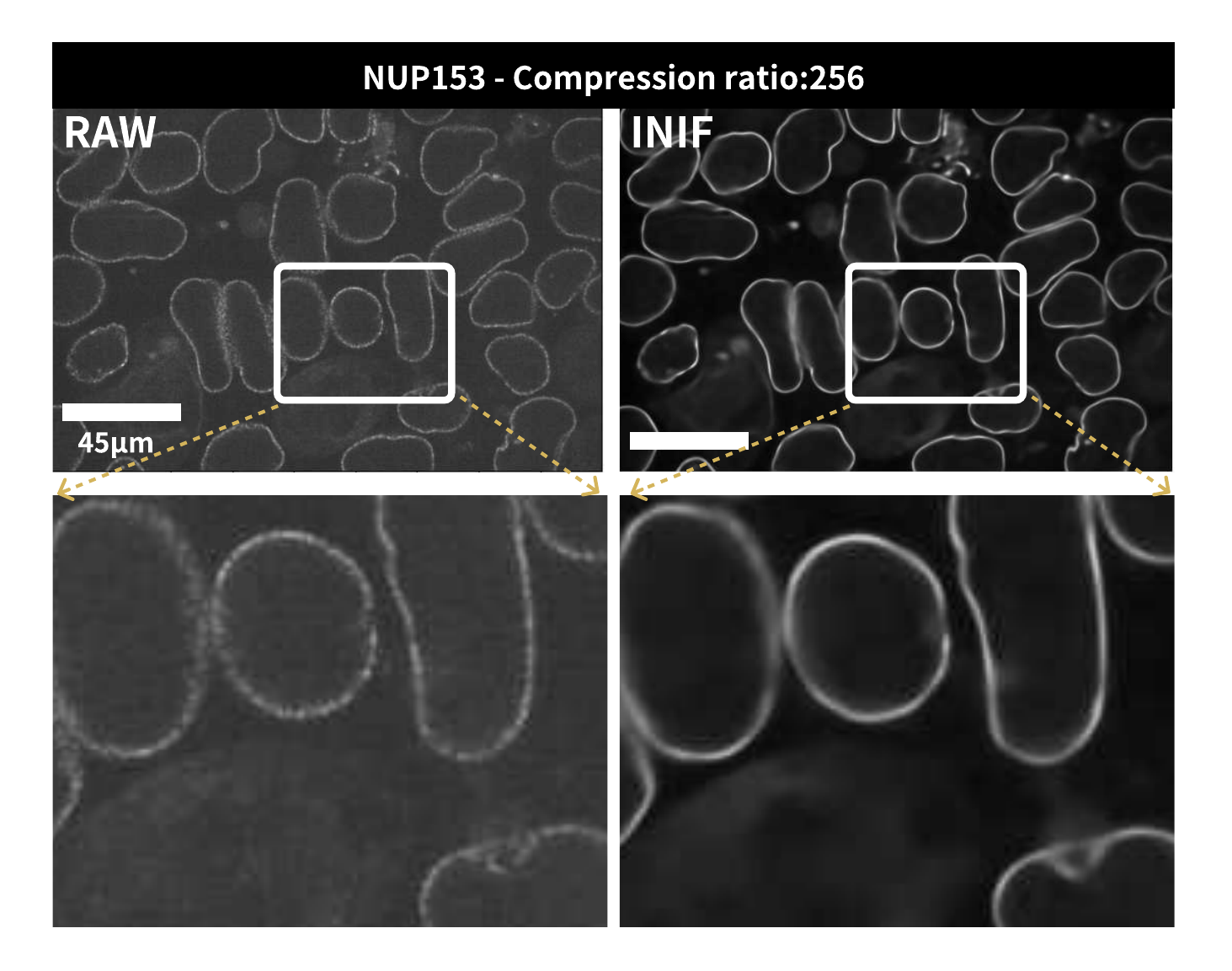}
    \caption{\textbf{Over-smooth artifacts produced by INR-based compression.} The protein NUP153 is associated with nuclear pores. The structural analysis of raw images indicates that it should exhibit discrete dots arranged around the nuclei. However, INR networks have a tendency to acquire a continuous representation.}
    \label{supfig6}
\end{figure}
Although we consider the INR method as a promising compressor due to its flexibility and differentiable nature, there are some observed limitations that need clarification. One major drawback of INR-based methods is their computational expense. Compared to traditional codecs, these methods require iterative optimization by gradient descent using neural networks, resulting in significantly slower compression speeds on the same hardware. In order to address this issue, we propose two effective designs in INIF that might be a feasible solution for future discussions. Firstly, we suggest utilizing a learned optimizer which enables dynamic step length prediction for update values. This approach involves using a hypernetwork to directly predict the final weight of the network (i.e., learned optimizer with one-step optimization), thereby boosting convergence speed. Secondly, we explore leveraging hand-crafted codecs' capabilities by transforming INIF into an adapter for such codecs, allowing mutual improvement for both methods.

The second drawback lies in the tendency of neural networks to learn continuous functions, posing a challenge in preventing excessive smoothing even with the INIF workflow. As illustrated in Figure~\ref{supfig6}, the compression of nuclear pore-related images yields unsatisfactory results as it deviates from the actual structure. We address this issue by introducing application-specific guidance in Section~\ref{sec2.2}. Although completely resolving the problem of over-smoothing remains elusive, we can still utilize guidance such as segmentation loss to prioritize the preservation of crucial information.

\end{appendices}
\end{document}